%% file: example_paper.tex
\documentclass{article}

\usepackage{microtype}
\usepackage{graphicx}
\usepackage{subcaption}
\usepackage{booktabs} 

\usepackage{hyperref}

\usepackage[preprint]{icml2027}

\usepackage{amsmath}
\usepackage{amssymb}
\usepackage{mathtools}
\usepackage{amsthm}

\usepackage[capitalize,noabbrev]{cleveref}

\theoremstyle{plain}

\theoremstyle{definition}

\theoremstyle{remark}

\usepackage[textsize=tiny]{todonotes}

\usepackage{booktabs,multirow,graphicx,adjustbox}
\icmltitlerunning{LoopFill: Pipelined Multi-Token Speculative Decoding for Looped Transformers}

\begin{document}

\twocolumn[
  \icmltitle{LoopFill: Pipelined Multi-Token Speculative Decoding for Looped Transformers}



  \icmlsetsymbol{equal}{*}

  \begin{icmlauthorlist}
    \icmlauthor{Siqi Zhu}{uiuc}
    \icmlauthor{Suozhi Huang}{pton}
    \icmlauthor{Yi Zhao}{uiuc}
    \icmlauthor{Jiaxuan You}{uiuc}
  \end{icmlauthorlist}

  \icmlaffiliation{uiuc}{University of Illinois Urbana-Champaign}
  \icmlaffiliation{pton}{Princeton University}

  \icmlcorrespondingauthor{Siqi Zhu}{siqizhu4@illinois.edu}

  \icmlkeywords{Machine Learning, ICML}

  \vskip 0.3in
]



\printAffiliationsAndNotice{}  

\begin{abstract}

Looped Transformers improve parameter efficiency by repeatedly applying the same Transformer layers. During autoregressive decoding, however, generating each token requires several sequential passes through these layers, leading to high decoding latency. Pipelined self-speculative methods use shallow-loop predictions to process tokens at different loops in the same batch, but each token must reach the drafting depth before the next can enter the pipeline. We introduce LoopFill, a training-free speculative decoding algorithm for looped LLM that combines multi-token suffix drafts with shallow-loop predictions in a single draft tree. Both drafters contribute to
one tree, so a suffix lookup can immediately extend a shallow prediction
and several future tokens can enter the pipeline together. The shared
block computes the highest-scoring draft candidates at different loop depths in one
batch, and full-depth verification preserves the model's autoregressive
output distribution. Across seven benchmarks, LoopFill's average speedup over
greedy autoregressive decoding is $5.85\times$ on Ouro-2.6B and
$5.13\times$ on Huginn-3.5B. The speedups exceed those of LoopAhead, the strongest baseline, by \(40\%\) on Ouro-2.6B and \(25\%\) on Huginn-3.5B. 


\end{abstract}

\input{sections/01_introduction}

\input{sections/02_preliminary}

\input{sections/03_method}

\input{sections/035_theory.tex}

\input{sections/04_experiments}

\input{sections/05_relatedwork}

\input{sections/06_conclusion}

\bibliography{example_paper}
\bibliographystyle{icml2027}

\newpage
\appendix
\onecolumn
\input{sections/07_appendix}
\end{document}

%% file: sections/01_introduction.tex
\section{Introduction}
\label{sec:intro}

Looped Transformer is a parameter-efficient architecture for reasoning language models~\cite{dehghani2019universaltransformers,
geiping2025scalingtesttimecomputelatent,zhu2026scalinglatentreasoninglooped,
bae2025mixtureofrecursionslearningdynamicrecursive}. They usually consist of a
prelude, a weight-shared block applied $T$ times, and a coda
(Figure~\ref{fig:loop-architecture}). Iterating the block refines the
hidden state without emitting tokens, favoring latent
reasoning~\cite{saunshi2025reasoninglatentthoughtspower}: Huginn-3.5B
improves as its test-time recurrence grows~\cite{geiping2025scalingtesttimecomputelatent},
and the 1.4B and 2.6B Ouro models match larger standard
LLMs~\cite{zhu2026scalinglatentreasoninglooped}. The price is slow
decoding. Each token requires $T$ passes through the block, and at small batch sizes each pass is
memory-bound~\cite{pope2022efficientlyscalingtransformerinference},
leaving most of the compute idle.

\begin{figure}[h]
  \centering
  \includegraphics[width=0.7\columnwidth]{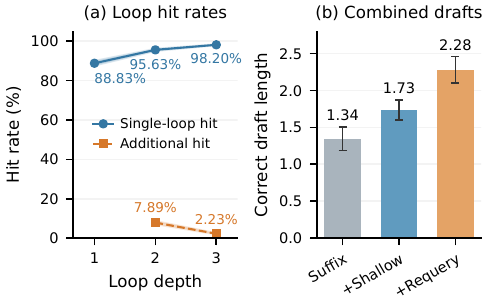}
  \caption{\textbf{Loop-$d$ and suffix-tree drafts on Ouro-2.6B.} (a) Blue: how often the loop-$d$ draft equals the
    full-depth output. Orange: additional loop 2 hits where loop 1 misses, and additional loop 3 hits where loops 1 and 2 both
    miss. (b) Mean length of the correct draft for suffix drafts alone, adding loop-1 draft
    (+Shallow), and also adding suffix draft after loop-1 draft (+Requery)}
  \label{fig:loopfill-preliminary}
\end{figure}

LoopSpec~\cite{cho2026loopspecpipelinedselfspeculativedecoding} and
WaveFront Decoding~\cite{ha2026wavefrontdecodingparallelizedselfspeculative}
turn this recurrence into a speculative
decoding pipeline. After $d<T$ loops, the
output head reads a draft token from the hidden state, and the next token
starts from this draft while the current one continues to depth $T$ to
verify it. Since the loops share weights, positions at different depths
run in one batched pass. This pipeline relies on an inductive bias of
looped models, which are trained to predict at varying
depths~\cite{geiping2025scalingtesttimecomputelatent,zhu2026scalinglatentreasoninglooped}:
shallow-loop predictions agree closely with full-depth ones. On
Ouro-2.6B, the top-1 token after one of $T=4$ loops matches the
full-depth greedy token at 88.8\% of positions, and after two loops at
95.6\% (Figure~\ref{fig:loopfill-preliminary}(a)).

However, these methods draft only along the recurrent depth. Each
shallow prediction proposes one token, so a new token starts at most
once every $d$ loop steps and the speedup is bounded by $T/d$. A rejected draft further forces a restart.
These bubbles leave the pipeline underfilled, while the sequence
dimension, i.e., starting several future tokens together, stays unused.
LoopAhead~\cite{liu2026loopaheadspeculatingaheadfaster} adds
multi-token prediction using trainable
heads
(Table~\ref{tab:suffix-loop-design}). We find that a training-free
drafter suffices. A suffix tree over the prompt and previous
outputs~\cite{oliaro2025suffixdecodingextremespeculativedecoding}
retrieves multi-token continuations without model computation. Combined
with shallow drafts, it extends the correct draft path on Ouro-2.6B from
1.34 to 1.73 tokens, and to 2.28 tokens when the suffix tree is also
queried at the loop-1 draft (Figure~\ref{fig:loopfill-preliminary}(b)).
The pairing is natural: shallow-loop drafting shortens the wait between
the starts of adjacent tokens, while suffix-tree drafting proposes
several future tokens without waiting for any loop.

\begin{table}[t]
  \centering
  \caption{\textbf{Comparison with existing speculative decoding methods
from four perspectives}. LoopFill
  combines multi-token suffix drafts with shallow-loop draft from the
  target model in a draft tree without training.
    }
  \label{tab:suffix-loop-design}
  \scriptsize
  \setlength{\tabcolsep}{0.8pt}
  \renewcommand{\arraystretch}{1}
  \def\sdyes{{\color[HTML]{228B22}\usefont{U}{pzd}{m}{n}\char51}}
  \def\sdno{{\color[HTML]{FF0000}\usefont{U}{pzd}{m}{n}\char55}}
  \begin{tabular*}{\columnwidth}{@{}l@{\extracolsep{\fill}}cccc@{}}
    \hline
    Method & Data & Model & Multi-token & Training-free \\
    \hline
    \multicolumn{5}{@{}l}{\textit{Data-driven drafting}} \\
    SuffixDecoding \cite{oliaro2025suffixdecodingextremespeculativedecoding}
      & \sdyes & \sdno  & \sdyes & \sdyes \\
    \quad + EAGLE-3 \cite{li2025eagle3scalinginferenceacceleration}
      & \sdyes & \sdyes & \sdyes & \sdno  \\
    \hline
    \multicolumn{5}{@{}l}{\textit{Model-driven drafting}} \\
    LoopSpec \cite{cho2026loopspecpipelinedselfspeculativedecoding}
      & \sdno  & \sdyes & \sdno  & \sdyes \\
    WaveFront \cite{ha2026wavefrontdecodingparallelizedselfspeculative}
      & \sdno  & \sdyes & \sdno  & \sdyes \\
    LoopAhead \cite{liu2026loopaheadspeculatingaheadfaster}
      & \sdno  & \sdyes & \sdyes & \sdno  \\
    DFlash \cite{chen2026dflashblockdiffusionflash}
      & \sdno  & \sdyes & \sdyes  & \sdno \\
    \hline
    \rlap{{\setlength{\fboxsep}{0pt}%
      \colorbox[HTML]{E6F7FD}{\makebox[\columnwidth][l]{\strut}}}}%
    \textbf{LoopFill}
      & \sdyes & \sdyes & \sdyes & \sdyes \\
    \hline
  \end{tabular*}
\end{table}

We propose LoopFill, which coordinates the two drafters in a unified
draft tree (Figure~\ref{fig:loopfill-overview}). Both drafters attach
candidates to the same tree, so a suffix continuation can follow a
shallow draft and vice versa, and a token proposed by both shares one
node. Every draft node occupies a row of the batched block at every
loop, so discarded drafts take compute from other requests. LoopFill
therefore scores each node by the product of shallow-prediction
probabilities and suffix frequencies along its path, computes only the
top-$N$ nodes in each loop step, and prunes branches that a later
shallow prediction rules out. When the root completes $T$ loops, the
matching subtree is kept together with its completed loops. Drafts
decide only which positions start early, never which token is emitted,
so LoopFill preserves the output of autoregressive decoding. The best per-request draft
budget depends on hardware and workload: On Huginn-3.5B, increasing the draft node budget from $N=16$ to $N=64$
raises the decoding speed by 19.5\% at $B=1$ and reduces
end-to-end throughput by 43\% at $B=16$ (Figure~\ref{fig:loopfill-load-budget}).

Our contributions are as follows:
\begin{itemize}
  \item We propose LoopFill, a training-free, pipelined multi-token
    speculative decoding method for looped language models. It combines
    a data-driven suffix-tree drafter with the model's own shallow-loop
    drafts to fill the bubbles of the loop pipeline, while preserving
    autoregressive outputs.
  \item We design a dual-drafter draft tree that shares computation,
    verification, and reuse between the two drafters, with path-score
    pruning under a node budget.
  \item On seven benchmarks, LoopFill achieves mean speedups of
$5.13$-$5.85\times$ over greedy autoregressive decoding on Ouro-1.4B,
Ouro-2.6B, and Huginn-3.5B. The speedups are $1.25$-$1.42\times$
those of the strongest baseline. LoopFill has $2.50$-$2.81\times$ the end-to-end
throughput of autoregressive decoding.
\end{itemize}

%% file: sections/02_preliminary.tex
\section{Preliminaries}
\label{sec:prelim}

\begin{figure*}[t]
  \centering
    \begin{minipage}[c]{0.16\textwidth}
      \centering
      \includegraphics[width=\linewidth]{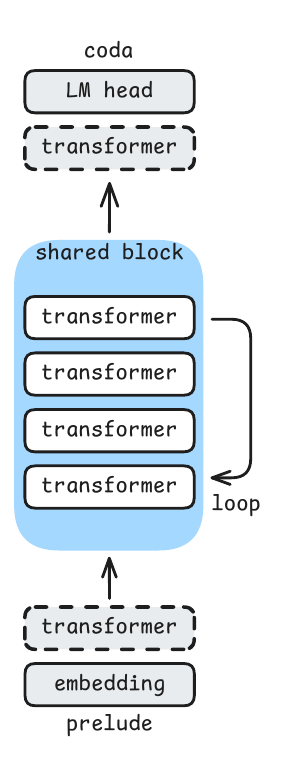}
      \par\smallskip
      \textbf{(a)}
    \end{minipage}
    \hspace{0.02\textwidth}
    \begin{minipage}[c]{0.8\textwidth}
      \centering
      \includegraphics[width=\linewidth]{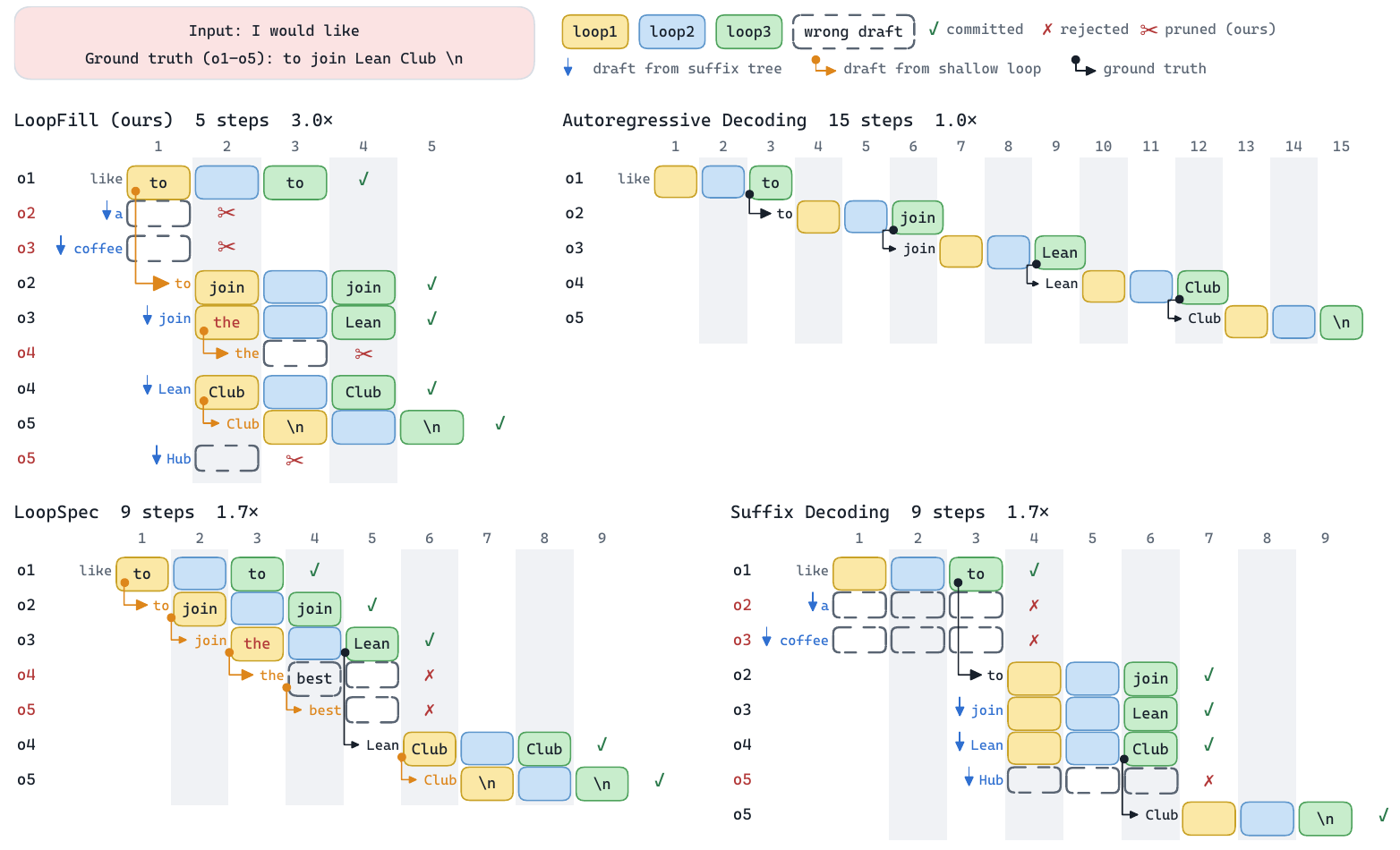}
      \par\smallskip
      \textbf{(b)}
    \end{minipage} 
  \caption{
    \textbf{(a) Looped LLM architecture.} In Ouro \cite{zhu2026scalinglatentreasoninglooped},
the entire Transformer stack is recurrent, with the embedding and LM head
outside the loop. Huginn \cite{geiping2025scalingtesttimecomputelatent}
places the recurrent block between prelude and coda blocks with Transformer layers.
    \textbf{(b) Comparing four decoding algorithms}
    (3 loops).
    Row $o_k$ is the position that predicts
    the $k$-th output token, with its input token on the left,
    and each column is one loop step.
    Autoregressive decoding starts a token after
    the previous one finishes.
    LoopSpec starts new token after one loop step, but the wrong draft
    (e.g. \texttt{the}) forces a restart.
    SuffixDecoding verifies multi-token drafts but waits
    for all three loops in every round.
    LoopFill (Section~\ref{sec:loopfill}) takes 5 loop steps to finish, achieving $3\times$ speedup.
  }
  \label{fig:four-decoders}

  \begingroup
    \makeatletter
    \edef\@currentlabel{\thefigure(a)}
    \label{fig:loop-architecture}
    \edef\@currentlabel{\thefigure(b)}
    \label{fig:four-decoders-panel}
  \endgroup
\end{figure*}



\paragraph{Looped language models.}
As shown in Figure~\ref{fig:loop-architecture}, a looped language
model encodes its input with a prelude $\mathcal{P}$, applies
a shared stack of Transformer layers $f$ for $T$ iterations, and
predicts the next token with a coda $\mathcal{C}$~\citep{cho2026loopspecpipelinedselfspeculativedecoding,liu2026loopaheadspeculatingaheadfaster}.
Given a prefix $x$ ending in $o_{i-1}$, the distribution for the
next token $o_i$ at depth $t$ is calculated as
\begin{equation}
\label{eq:loop-recurrence}
\begin{aligned}
  E &= \mathcal{P}(x), \\
  H^{(t)} &= f\!\left(H^{(t-1)}, E\right), \qquad t=1,\ldots,T, \\
  p_t(\cdot\mid x) &= \bigl[\mathcal{C}(H^{(t)})\bigr]_{\mathrm{last}}.
\end{aligned}
\end{equation}
Depending on the architecture, the prefix encoding $E$ initializes
the recurrent state $H^{(0)}$ \cite{zhu2026scalinglatentreasoninglooped} or serves as a fixed input to each
loop of $f$ \cite{geiping2025scalingtesttimecomputelatent}. The coda includes the LM head and softmax, and
$[\cdot]_{\mathrm{last}}$ selects the row for the last prefix token.
We denote the corresponding hidden state by
$h_i^{(t)}=[H^{(t)}]_{\mathrm{last}}$, indexing it by the token
$o_i$ it predicts. Standard autoregressive decoding uses the
full-depth output $o_i$ as input to predict $o_{i+1}$, so
computations for consecutive tokens start $T$ iterations apart.

\paragraph{Pipelined self-speculative decoding for looped LMs.}
Pipelined decoding uses a draft for $o_i$ to begin predicting
$o_{i+1}$ before that draft is verified. SPEED~\citep{hooper2024speedspeculativepipelinedexecution}
uses cyclic weight sharing to overlap these computations and restarts
dependent work when a deeper layer revises a prediction.
LoopSpec~\citep{cho2026loopspecpipelinedselfspeculativedecoding}
obtains its draft $\hat{o}_i$ directly from $p_d(\cdot\mid x) =\bigl[\mathcal{C}(H^{(d)})\bigr]_{\mathrm{last}}$ at
a shallow depth $1\leq d<T$, without additional training. The looped model
then processes $x$ and the prefix extended by $\hat{o}_i$ in parallel
(Figure~\ref{fig:four-decoders-panel}). Each loop step advances both computations by
one loop in a batched pass through $f$. A deeper loop can
propose an alternative for $o_i$ if the model's confidence in
$\hat{o}_i$ falls. At depth $T$, LoopSpec verifies the candidates
using $p_T(\cdot\mid x)$ and reuses the work on the accepted branch,
restarting from a replacement token only if all draft candidates are rejected.

\paragraph{Model-free drafting for speculative decoding.}
Model-free speculative decoding generates draft tokens from
previous text or decoding history without an auxiliary
draft model. LLMA and Prompt Lookup Decoding reuse continuations
from matching prompt spans~\cite{yang2023inferencereferencelosslessacceleration,saxena2023prompt}, ANPD and Token Recycling use $n$-grams and
cached token~\cite{ou-etal-2024-lossless,luo-etal-2025-turning}. 
SuffixDecoding~\cite{oliaro2025suffixdecodingextremespeculativedecoding}
extends this approach with suffix trees built from prompts and
historical outputs, adaptively constructing draft trees. The target LLM verifies the draft tree in parallel and
accepts the draft prefix with one
additional token by full-depth decoding.
In a looped model, each verification round takes $T$ loop steps,
and all token computations in the round start together.
The next round begins after the current round finishes.
In the SuffixDecoding panel of
Figure~\ref{fig:four-decoders-panel},
the three rounds commit 1, 3, and 1 tokens.
With $T=3$, generating these five tokens takes nine loop steps.

%% file: sections/03_method.tex
\section{LoopFill Decoding}
\label{sec:loopfill}

LoopFill maintains one draft tree $\mathcal D$ per request. The root
represents the verified prefix $x$. Each child represents its parent's
prefix extended by one candidate token. For example, if the root represents
\texttt{I would like}, its child labeled \texttt{to} represents
\texttt{I would like to}. While the root predicts the token after
\texttt{like}, that child can already begin predicting the token after
\texttt{to}. Suffix retrieval and shallow-loop predictions add
children to the same tree. If both propose the same token under the same
parent, LoopFill creates a single child and computes its state once.


\begin{figure}[t]
  \centering
  \includegraphics[width=0.8\columnwidth]{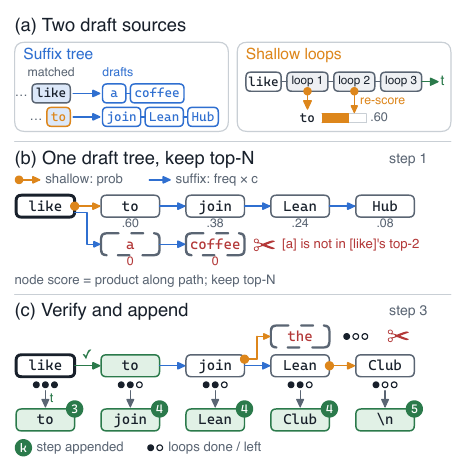}
  \caption{\textbf{LoopFill decoding.}
    (a) Suffix retrieval extends the shallow draft \texttt{to}.
    (b) After step~1, \texttt{a} falls outside the shallow top-2 after
    \texttt{like}, so both \texttt{a} and its child \texttt{coffee}
    have a path score of zero.
    (c) The tree after step~3. Dots show how many loops each node has completed toward predicting the token after its prefix. The numbered badges mark the loop step at which each output token is appended.}
  \label{fig:loopfill-overview}
\end{figure}

\paragraph{Proposing draft tokens.}
LoopFill searches two suffix trees \cite{oliaro2025suffixdecodingextremespeculativedecoding}. The local tree indexes the current
request's prompt and the tokens generated so far. The global tree
indexes the outputs of completed requests.
For the prefix \texttt{I would like}, LoopFill looks
in both trees for \texttt{like}, \texttt{would like}, and
\texttt{I would like}. If \texttt{would like} was followed by
\texttt{to} in a previous request, LoopFill adds \texttt{to} as
a candidate next token. In general, LoopFill checks every suffix up to 64 tokens. We use $p$ to denote the
length of the suffix being matched.

For each matching suffix, LoopFill builds a small tree
from the tokens that followed that suffix. It repeatedly adds the
highest-scoring token up to $\min(Q,p)$ tokens
in total. The score of an extension is the product of the
next-token frequencies along its path. For example, the score for
\texttt{join Lean} multiplies the fraction of matches followed by
\texttt{join} by the fraction of those cases followed in turn by
\texttt{Lean}. Across all suffix lengths and both suffix trees, LoopFill selects
the candidate subtree with the largest sum of node scores. It merges
this subtree into $\mathcal{D}$.


When a node in $\mathcal{D}$ reaches a designated shallow drafting depth $d<T$,
LoopFill applies the LM head to its hidden state and proposes its top-$k$ tokens as children.
 Figure~\ref{fig:loopfill-overview} illustrates how the
two sources interact. At the root, suffix retrieval proposes
\texttt{a coffee} and the shallow draft proposes \texttt{to}. Then, a suffix
lookup for \texttt{like to} proposes \texttt{join Lean Hub}. When the node ending in \texttt{Lean} reaches a
drafting depth, its shallow prediction adds \texttt{Club} alongside \texttt{Hub}.

\paragraph{Selecting drafts for computation.}
LoopFill ranks nodes from both draft sources to decide which prefixes to
advance in the next loop step. When suffix retrieval creates a new child
$v$, it assigns the edge from its parent the initial weight
\begin{equation}
  w(v)=\min\{1,c\,q(v)\},
  \label{eq:edge-score}
\end{equation}
where \(q(v)\) is the fraction of matches at the parent node that continue with the child $v$, and $c$ is a scaling factor.
Retrieval leaves existing weights in place. When a node makes a shallow
prediction at loop $d$, it updates the weights of all its children. The top $k$
tokens receive their model probabilities, and the remaining children
receive zero.

LoopFill ranks each draft node by the product of the weights along
its path from the root,
\begin{equation}
  s(\mathrm{root})=1,\qquad
  s(v)=s(\mathrm{parent}(v))\,w(v).
  \label{eq:path-score}
\end{equation}
A child with zero weight and all its descendants have a path score of
zero. They rank last but can still be kept if the decoding budget allows.

After a shallow prediction or a root change, let $\mathcal{V}$ contain
all non-root nodes in $\mathcal{D}$.
LoopFill ranks nodes by path score, regardless of
which source proposed them. It keeps the root and the top $N$ draft nodes for each request,
\begin{equation}
  \mathcal{D}\gets\{\mathrm{root}\}
  \cup\operatorname{Top}_N(\mathcal{V},s).
  \label{eq:top-n}
\end{equation}
A parent always ranks ahead of its children. Therefore, if we keep a node, it also keeps every node on the path back to the root. LoopFill retrieves suffix drafts at the root when decoding begins
and whenever the root changes. It ranks these drafts, any new
shallow drafts, and the existing draft nodes together, keeping up
to $N$ nodes. It then uses new shallow draft as a
prefix for another suffix draft. For example, if the shallow draft
\texttt{to} is kept, LoopFill looks up continuations of
\texttt{I would like to} and adds them below \texttt{to}. It ranks
the expanded tree again and keeps the top $N$ draft nodes.

Each loop step advances node by one loop. A batched
pass of the loop block processes nodes from different requests and
at different loop depths. Pruned nodes lose
their completed work. Each request uses at most $N+1$ slots per
step, including the root. In our default setting, we use $N=16$, $top k=2$, and $Q=4$ for max throughput.

\paragraph{Verification and reuse.}
Once the root finishes $T$-loop verification, its next-token distribution
$p_T(\cdot\mid x)$ determines the output token,
\begin{equation}
  o_i =
  \begin{cases}
    \arg\max_a p_T(a\mid x), & \text{greedy decoding},\\
    F_x^{-1}(\xi_i),         & \text{sampling}.
  \end{cases}
  \label{eq:verify}
\end{equation}
Here $F_x$ is the cumulative distribution at a sampling
temperature. LoopFill draws $\xi_i\sim\mathcal{U}(0,1)$ once per
output position.

LoopFill appends $o_i$ to the output and updates the local suffix
tree. It then checks the children of the root in the draft tree
$\mathcal{D}$. If a child matches $o_i$, LoopFill promotes it to
the root, preserving its completed loops and descendants, and
discards the other branches. If no child matches, LoopFill starts
a new root at loop zero for the extended prefix. A new root that
has already finished $T$ loops can immediately choose the following
token. In Figure~\ref{fig:loopfill-overview}c, \texttt{like} selects
\texttt{to} at step~3. At step~4, \texttt{to}, \texttt{join}, and
\texttt{Lean} finish together, so LoopFill appends \texttt{join Lean Club}
in one step.

%% file: sections/035_theory.tex
\section{Theoretical Analysis}
\label{sec:theory}

A suffix draft lets LoopFill process several consecutive tokens in
parallel. If the suffix draft misses the ground-truth token at position $i$: $o_i$, a shallow draft can still supply $o_i$ for computing $o_{i+1}$ before verification of $o_i$ finishes.
We assume one request during serving and enough capacity to
process every draft immediately. Each draft advances
one loop per step. Verification requires $T$ loops.

For a request that generates $n$ output tokens in $K$ sequential
decoding loop steps after prefill, we define the
\textbf{loop-normalized mean accepted length} as
\begin{equation}
  \gamma = \frac{nT}{K}.
  \label{eq:loop-normalized-length}
\end{equation}
 $\gamma$ measures how many output tokens are generated per
$T$ sequential loop steps. Standard autoregressive decoding takes
$K=nT$ steps, giving $\gamma=1$.

Let $t_i$ be the loop step when the model starts predicting the
$i$th such token $o_i$ from the correct prefix. Set $t_1=0$ and
let $\delta_i=t_i-t_{i-1}$ denote the delay between consecutive
starts for $i\ge 2$. The final token is verified at $t_n+T$, so
$K=t_n+T=T+\sum_{i=2}^{n}\delta_i$. If the mean delay converges
to $\Delta>0$ as $n\to\infty$, then $K/n\to\Delta$ and
$\gamma\to T/\Delta$. We use this limiting value in the analysis
below.

\noindent\textbf{Proposition (Shallow drafts on suffix misses).}
Suppose the suffix draft is available at $t_i$ and the shallow draft
at $t_i+d$, where $0<d<T$. A draft hits if its next-token candidates
include $o_i$. Let $E_s$ and $E_m$ denote suffix and shallow hits,
with limiting frequencies $p_s=\Pr(E_s)<1$ and $p_m=\Pr(E_m)$ along
the output sequence. The conditional rate $\alpha=\Pr(E_m\mid E_s^c)$
is the shallow hit rate on suffix misses. Then
\begin{equation}
  \gamma_{\mathrm{LF}}
  =\underbrace{\frac{1}{1-p_s}}_{\gamma_{\mathrm{suffix}}}
   \frac{T}{\alpha d+(1-\alpha)T}.
  \label{eq:combined}
\end{equation}

\emph{Proof.}
On a suffix hit, computation for $o_{i+1}$ starts at $t_i$, so
$\delta_{i+1}=0$. On a suffix miss, the delay is $d$ if the shallow
draft hits and $T$ otherwise. On average, the correct-prediction delay for LoopFill is
$\Delta_{\mathrm{LF}}=(1-p_s)[\alpha d+(1-\alpha)T]$.
For suffix decoding, $\Delta_{\mathrm{suffix}}=(1-p_s)T$.
Substituting both delays into $\gamma=T/\Delta$ produces
Eq.~\eqref{eq:combined}. If each drafter proposes the same candidates
as when used alone, LoopFill has the smaller of the two delays
for each token.
Therefore
$\gamma_{\mathrm{LF}}\geq
\max\{\gamma_{\mathrm{suffix}},\gamma_{\mathrm{shallow}}\}$.

The second factor in Eq.~\eqref{eq:combined} measures the gain from
shallow drafting on suffix misses. A shallow hit on a suffix miss
saves $T-d$ steps, giving an average saving of $(1-p_s)\alpha(T-d)$
steps per output token over suffix decoding. The standalone
shallow hit rate $p_m$ also counts tokens that both drafts predict
correctly. When $\alpha<p_m$, multiplying the two standalone gains overestimates $\gamma_{\mathrm{LF}}$.

With shallow drafts alone, consecutive token computations start at
least $d$ steps apart, so $\gamma_{\mathrm{shallow}}\leq T/d$.
Relative to this $d$-step interval, a suffix hit saves $d$ steps.
If both drafts miss, the extra delay is $T-d$ steps. Comparing the
mean saving with the mean extra delay gives
\begin{equation}
  \gamma_{\mathrm{LF}}>T/d
  \quad\Longleftrightarrow\quad
  d\,p_s>(T-d)(1-p_s)(1-\alpha).
  \label{eq:pipeline-ceiling}
\end{equation}
On our runtime traces for Ouro-2.6B ($T=4$, top-1 shallow draft
at $d=1$), $p_s=0.510$ and $\alpha=0.796<p_m=0.888$.
Equation~\eqref{eq:combined} predicts $\gamma_{\mathrm{LF}}\approx5.06$,
above the shallow-only ceiling of $4$.

For multiple shallow-draft depths, let
$0=d_0<d_1<\cdots<d_J=T$ be the possible delays. Let
$q_j=\Pr(\delta_{i+1}=d_j)$ be the fraction of output tokens first
available as input to the next prediction at delay $d_j$. Tokens that
all drafts miss contribute to $q_J$. Then
\begin{equation}
  \Delta=\sum_{j=0}^{J}q_jd_j,
  \qquad \gamma=\frac{T}{\Delta}.
  \label{eq:multilevel-delay}
\end{equation}
For example, suppose a shallow draft first proposes $o_i$ at depth
$d_j$. If a suffix lookup using the prefix ending in $o_i$ also
proposes $o_{i+1}$ in that step, then $\delta_{i+1}=d_j$ and
$\delta_{i+2}=0$.

With a finite draft budget $N$, a suffix draft can displace a
correct shallow candidate. Shallow-only decoding could then start
computing the next token earlier than LoopFill. Under concurrent serving,
extra draft nodes consume loop passes that could serve other requests.
Figure~\ref{fig:loopfill-load-budget} shows how $N$ affects $\gamma$,
loop passes per output token, and throughput.

%% file: sections/04_experiments.tex
\section{Experiments}

\begin{table*}[ht]
\centering
\caption{
Decode latency speedup ($\eta$) and loop-normalized mean accepted length ($\gamma$) with Temperature = 0.0 \& 1.0 and concurrency = 1 on seven benchmarks. Mean values are arithmetic averages on the seven benchmarks.
Best results are highlighted in \textbf{bold}.
}
\label{tab:main_results}
\vspace{-1mm}
\begingroup
\footnotesize
\renewcommand{\arraystretch}{1}
\setlength{\tabcolsep}{3pt}
\begin{adjustbox}{max width=\textwidth}
\begin{tabular}{@{}lcl*{16}{c}@{}}
\toprule
\multirow{2}{*}{Model} & \multirow{2}{*}{Temp.} & \multirow{2}{*}{Method}
& \multicolumn{2}{c}{GSM8K}
& \multicolumn{2}{c}{MATH-500}
& \multicolumn{2}{c}{BBH}
& \multicolumn{2}{c}{HumanEval+}
& \multicolumn{2}{c}{MBPP+}
& \multicolumn{2}{c}{SpecBench}
& \multicolumn{2}{c}{Spider}
& \multicolumn{2}{c}{Mean} \\
& &
& $\eta$ & $\gamma$
& $\eta$ & $\gamma$
& $\eta$ & $\gamma$
& $\eta$ & $\gamma$
& $\eta$ & $\gamma$
& $\eta$ & $\gamma$
& $\eta$ & $\gamma$
& $\eta$ & $\gamma$ \\
\midrule
\multirow{14}{*}{Huginn-3.5B} & \multirow{7}{*}{0.0} & AR
& 1.00 & 1.00
& 1.00 & 1.00
& 1.00 & 1.00
& 1.00 & 1.00
& 1.00 & 1.00
& 1.00 & 1.00
& 1.00 & 1.00
& 1.00 & 1.00 \\
 &  & WaveFront
& 2.32 & 3.05
& 2.86 & 4.00
& 2.95 & 3.96
& 3.48 & 4.62
& 2.83 & 3.69
& 2.36 & 3.11
& 3.01 & 4.17
& 2.83 & 3.80 \\
 &  & LoopSpec
& 3.47 & 4.88
& 3.99 & 5.58
& 4.22 & 5.79
& 4.33 & 6.08
& 3.96 & 5.33
& 3.46 & 4.65
& 4.10 & 5.57
& 3.93 & 5.41 \\
 &  & SuffixDecoding
& 1.58 & 1.75
& 1.97 & 2.18
& 2.50 & 2.70
& 2.09 & 2.36
& 1.79 & 1.87
& 1.51 & 1.57
& 2.03 & 2.22
& 1.92 & 2.09 \\
 &  & DFlash
& 2.35 & 2.85
& 4.61 & 5.22
& 3.81 & 4.25
& 5.24 & 6.48
& 4.22 & 5.22
& 2.58 & 2.90
& 3.28 & 3.46
& 3.73 & 4.34 \\
 &  & LoopAhead
& 3.32 & 5.41
& 4.29 & 6.96
& 4.28 & 6.63
& 5.26 & 7.95
& 4.19 & 5.97
& 3.27 & 4.95
& 4.22 & 6.27
& 4.12 & 6.30 \\
 &  & \textbf{LoopFill }
& \textbf{4.35} & \textbf{6.50}
& \textbf{5.30} & \textbf{7.72}
& \textbf{6.01} & \textbf{8.76}
& \textbf{5.85} & \textbf{8.53}
& \textbf{4.92} & \textbf{7.15}
& \textbf{4.08} & \textbf{5.91}
& \textbf{5.40} & \textbf{7.61}
& \textbf{5.13} & \textbf{7.45} \\
\cmidrule(l){2-19}
 & \multirow{7}{*}{1.0} & AR
& 1.00 & 1.00
& 1.00 & 1.00
& 1.00 & 1.00
& 1.00 & 1.00
& 1.00 & 1.00
& 1.00 & 1.00
& 1.00 & 1.00
& 1.00 & 1.00 \\
 &  & WaveFront
& 1.58 & 2.08
& 1.78 & 2.46
& 2.08 & 2.71
& 1.74 & 2.52
& 1.84 & 2.44
& 1.50 & 1.93
& 1.75 & 2.34
& 1.75 & 2.36 \\
 &  & LoopSpec
& 1.82 & 2.67
& 2.15 & 3.04
& 2.36 & 3.41
& 2.53 & 3.17
& 2.13 & 3.00
& 1.65 & 2.23
& 1.97 & 2.68
& 2.09 & 2.88 \\
 &  & SuffixDecoding
& 1.21 & 1.36
& 1.48 & 1.61
& 1.80 & 1.93
& 1.49 & 1.52
& 1.38 & 1.47
& 1.20 & 1.26
& 1.26 & 1.49
& 1.40 & 1.52 \\
 &  & DFlash
& 1.82 & 1.99
& 2.61 & 2.80
& 2.55 & 2.87
& 2.78 & 2.84
& 2.47 & 2.81
& 1.93 & 1.89
& 2.12 & 2.15
& 2.33 & 2.48 \\
 &  & LoopAhead
& 2.13 & 3.20
& 2.43 & 3.71
& 2.83 & 4.25
& 2.70 & 3.75
& 2.50 & 3.66
& 1.79 & 2.58
& \textbf{2.18} & 3.30
& 2.37 & 3.49 \\
 &  & \textbf{LoopFill}
& \textbf{2.34} & \textbf{3.72}
& \textbf{2.68} & \textbf{4.41}
& \textbf{2.86} & \textbf{4.59}
& \textbf{2.85} & \textbf{4.42}
& \textbf{2.94} & \textbf{4.49}
& \textbf{2.39} & \textbf{3.74}
& 2.11 & \textbf{3.97}
& \textbf{2.60} & \textbf{4.19} \\
\midrule
\multirow{14}{*}{Ouro-2.6B} & \multirow{7}{*}{0.0} & AR
& 1.00 & 1.00
& 1.00 & 1.00
& 1.00 & 1.00
& 1.00 & 1.00
& 1.00 & 1.00
& 1.00 & 1.00
& 1.00 & 1.00
& 1.00 & 1.00 \\
 &  & WaveFront
& 3.11 & 3.32
& 2.86 & 3.06
& 3.37 & 3.51
& 3.18 & 3.47
& 3.17 & 3.46
& 2.79 & 2.94
& 2.28 & 3.08
& 2.96 & 3.26 \\
 &  & LoopSpec
& 3.23 & 3.59
& 3.10 & 3.45
& 3.47 & 3.73
& 3.33 & 3.68
& 3.33 & 3.66
& 3.09 & 3.34
& 2.98 & 3.28
& 3.22 & 3.53 \\
 &  & SuffixDecoding
& 1.79 & 1.91
& 1.44 & 1.61
& 2.62 & 2.75
& 2.08 & 2.31
& 1.74 & 1.90
& 1.60 & 1.74
& 1.69 & 2.11
& 1.85 & 2.05 \\
 &  & DFlash
& 3.50 & 3.89
& 3.36 & 3.81
& 3.61 & 4.01
& 3.94 & 4.43
& 4.47 & 5.00
& 2.57 & 2.87
& 2.35 & 2.69
& 3.40 & 3.81 \\
 &  & LoopAhead
& 4.92 & 6.01
& 4.14 & 5.08
& 4.42 & 5.26
& 4.58 & 5.70
& 4.76 & 5.73
& 3.62 & 4.38
& 2.88 & 3.89
& 4.19 & 5.15 \\
 &  & \textbf{LoopFill }
& \textbf{5.76} & \textbf{6.79}
& \textbf{4.77} & \textbf{5.70}
& \textbf{8.23} & \textbf{9.38}
& \textbf{6.99} & \textbf{8.25}
& \textbf{5.70} & \textbf{6.72}
& \textbf{4.84} & \textbf{5.71}
& \textbf{4.64} & \textbf{6.23}
& \textbf{5.85} & \textbf{6.97} \\
\cmidrule(l){2-19}
 & \multirow{7}{*}{1.0} & AR
& 1.00 & 1.00
& 1.00 & 1.00
& 1.00 & 1.00
& 1.00 & 1.00
& 1.00 & 1.00
& 1.00 & 1.00
& 1.00 & 1.00
& 1.00 & 1.00 \\
 &  & WaveFront
& 2.41 & 2.55
& 2.35 & 2.55
& 2.60 & 2.75
& 2.36 & 2.50
& 2.37 & 2.51
& 1.97 & 2.07
& 2.08 & 2.28
& 2.30 & 2.46 \\
 &  & LoopSpec
& 2.63 & 2.70
& 2.43 & 2.73
& 2.70 & 2.83
& 2.35 & 2.68
& 2.40 & 2.65
& 2.09 & 2.18
& 2.07 & 2.44
& 2.38 & 2.60 \\
 &  & SuffixDecoding
& 1.49 & 1.66
& 1.37 & 1.56
& 1.97 & 2.08
& 1.39 & 1.58
& 1.50 & 1.57
& 1.24 & 1.39
& 1.21 & 1.57
& 1.45 & 1.63 \\
 &  & DFlash
& 2.29 & 2.75
& 2.44 & 2.83
& 2.77 & 3.02
& 2.53 & 3.00
& 2.55 & 2.97
& 1.98 & 2.11
& 1.63 & 2.10
& 2.31 & 2.68 \\
 &  & LoopAhead
& \textbf{3.83} & \textbf{4.51}
& 3.43 & \textbf{4.29}
& 3.33 & 3.98
& 3.11 & \textbf{4.29}
& 3.37 & \textbf{4.30}
& \textbf{2.71} & \textbf{3.29}
& 2.98 & 3.43
& 3.25 & 4.01 \\
 &  & \textbf{LoopFill }
& 3.65 & 4.15
& \textbf{3.47} & 4.18
& \textbf{4.45} & \textbf{5.12}
& \textbf{3.54} & 4.27
& \textbf{3.60} & 4.13
& 2.52 & 3.04
& \textbf{3.27} & \textbf{3.62}
& \textbf{3.50} & \textbf{4.07} \\
\bottomrule
\end{tabular}
\end{adjustbox}
\endgroup
\vspace{-1mm}
\end{table*}

\begin{table*}[ht]
\centering
\caption{
SpecBench end-to-end output throughput at different concurrency levels at decoding temperature $1.0$.
Each cell gives output tokens/s and the throughput ratio to AR at the same
concurrency. Best results are highlighted in bold.
}
\label{tab:specbench_throughput}
\resizebox{0.80\textwidth}{!}{%
\begin{tabular}{llccccc}
\toprule
\textbf{Model} & \textbf{Method}
& $B=1$ & $B=2$ & $B=4$ & $B=8$ & $B=16$ \\

\midrule

\multirow{7}{*}{Huginn-3.5B}
& AR
& 5.89 / 1.00$\times$
& 11.11 / 1.00$\times$
& 19.00 / 1.00$\times$
& 32.99 / 1.00$\times$
& 48.34 / 1.00$\times$ \\

\cmidrule(l){2-7}

& WaveFront
& 14.92 / 2.53$\times$
& 24.56 / 2.21$\times$
& 39.76 / 2.09$\times$
& 60.81 / 1.84$\times$
& 80.44 / 1.66$\times$ \\

& LoopSpec
& 19.65 / 3.34$\times$
& 33.89 / 3.05$\times$
& 51.03 / 2.69$\times$
& 74.24 / 2.25$\times$
& 89.65 / 1.85$\times$ \\

& SuffixDecoding
& 9.45 / 1.60$\times$
& 17.34 / 1.56$\times$
& 30.17 / 1.59$\times$
& 52.01 / 1.58$\times$
& 74.45 / 1.54$\times$ \\

& DFlash
& 18.36 / 3.12$\times$
& 33.18 / 2.99$\times$
& 57.47 / 3.03$\times$
& 88.18 / 2.67$\times$
& 117.52 / 2.43$\times$ \\

& LoopAhead
& 23.95 / 4.07$\times$
& 35.82 / 3.22$\times$
& 52.21 / 2.75$\times$
& 76.12 / 2.31$\times$
& 100.24 / 2.07$\times$ \\

& \textbf{LoopFill}
& \textbf{26.17 / 4.44$\times$}
& \textbf{40.73 / 3.67$\times$}
& \textbf{59.50 / 3.13$\times$}
& \textbf{94.08 / 2.85$\times$}
& \textbf{129.18 / 2.67$\times$} \\

\midrule

\multirow{7}{*}{Ouro-2.6B}
& AR
& 5.53 / 1.00$\times$
& 10.65 / 1.00$\times$
& 20.75 / 1.00$\times$
& 38.72 / 1.00$\times$
& 67.48 / 1.00$\times$ \\

\cmidrule(l){2-7}

& WaveFront
& 15.19 / 2.75$\times$
& 29.02 / 2.72$\times$
& 53.96 / 2.60$\times$
& 98.75 / 2.55$\times$
& 158.45 / 2.35$\times$ \\

& LoopSpec
& 16.77 / 3.03$\times$
& 32.02 / 3.01$\times$
& 58.78 / 2.83$\times$
& 104.06 / 2.69$\times$
& 166.52 / 2.47$\times$ \\

& SuffixDecoding
& 8.89 / 1.61$\times$
& 17.03 / 1.60$\times$
& 32.48 / 1.57$\times$
& 60.47 / 1.56$\times$
& 102.13 / 1.51$\times$ \\

& DFlash
& 16.17 / 2.93$\times$
& 28.85 / 2.71$\times$
& 55.83 / 2.69$\times$
& 96.72 / 2.50$\times$
& 153.48 / 2.27$\times$ \\

& LoopAhead
& 22.76 / 4.12$\times$
& 43.05 / 4.04$\times$
& 74.52 / 3.59$\times$
& 124.17 / 3.21$\times$
& 186.65 / 2.77$\times$ \\

& \textbf{LoopFill}
& \textbf{27.12 / 4.91$\times$}
& \textbf{51.17 / 4.80$\times$}
& \textbf{86.22 / 4.16$\times$}
& \textbf{140.94 / 3.64$\times$}
& \textbf{189.56 / 2.81$\times$} \\

\bottomrule
\end{tabular}%
}
\end{table*}

\paragraph{models and datasets.} 
We evaluate LoopFill on two representative looped language models, Ouro \cite{zhu2026scalinglatentreasoninglooped} and Huginn\cite{geiping2025scalingtesttimecomputelatent} on seven benchmarks: GSM8K \cite{gsm8k}, MATH-500 \cite{math}, BBH \cite{BBH}, HumanEval+ \cite{humaneval+}, MBPP+ \cite{humaneval+}, Spider \cite{yu2019spiderlargescalehumanlabeleddataset} and SpecBench \cite{specbench}. Our main experiments use Huginn-3.5B and Ouro-2.6B, with additional evaluations on Ouro-1.4B in Table \ref{tab:main_results_ouro14b} and \ref{tab:ouro14-specbench-throughput}. Table \ref{tab:main_results} shows results under decoding temperatures of 0.0 and 1.0, with max generation length of 512, Table \ref{tab:specbench_throughput} has max generation length of 1,024. To evaluate the DFlash and LoopAhead baselines, we additionally construct training sets covering math, coding, and reasoning. LoopAhead and DFlash training sets contain 9,025 and 50,000 examples, respectively. Details on dataset and configurations are in the appendix \ref{app:drafter-training}.

\paragraph{inference engine and metrics.}
We implement an inference
engine with continuous batching and chunked prefill for LoopFill and all baselines.
Each experiment uses one NVIDIA L20X GPU in BF16, further implementation details are in \ref{app:runtime_setup}.
We record decoding latency speedup $\eta$ against autoregressive decoding
and end-to-end output throughput (tokens/s).
Latency measures the decoding phase after prompt prefilling, throughput includes both prefill and decoding.
For each benchmark, we compute $\gamma$ using
Eq.~\eqref{eq:loop-normalized-length}. Here $n$ is the total
number of output tokens generated after prefill across all requests,
and $K$ is the sum of their decoding loop-step counts.

\subsection{Main Results}

We compare LoopFill with three types of baselines: \textbf{(1) Standard autoregressive decoding (AR)}. \textbf{(2) Single-token speculative methods for looped LM}. WaveFront~\cite{ha2026wavefrontdecodingparallelizedselfspeculative} and LoopSpec~\cite{cho2026loopspecpipelinedselfspeculativedecoding}. They typically use hidden states from early loops to draft the next token and begin processing it while the current token completes its remaining loops. \textbf{(3) Multi-token speculative methods}. SuffixDecoding~\cite{oliaro2025suffixdecodingextremespeculativedecoding} retrieves draft tokens from previous sequences, and DFlash~\cite{chen2026dflashblockdiffusionflash} trains a drafter to propose tokens. Neither method uses the loop pipeline. LoopAhead~\cite{liu2026loopaheadspeculatingaheadfaster} is the closest baseline, drafting multiple tokens within the loop pipeline. LoopAhead trains a drafter for each target model, whereas LoopFill combines suffix retrieval with the model's shallow loop predictions and requires no additional training.

Table~\ref{tab:main_results} shows the \textbf{decoding latency speedup over
AR ($\eta$)} and the \textbf{loop-normalized mean accepted length ($\gamma$)} for
LoopFill and the five speculative baselines. Under greedy decoding
(temperature~0), LoopFill is the fastest on all
seven benchmarks for both models. Its mean speedup $\eta$ is
$5.85\times$ on Ouro-2.6B and $5.13\times$ on Huginn-3.5B. Across
the two models, LoopFill's mean $\eta$ is $1.81$-$1.98\times$ that
of WaveFront, $1.31$-$1.82\times$ that of LoopSpec, and
$2.67$-$3.16\times$ that of SuffixDecoding. It exceeds DFlash's $\eta$ by 38-72\%. LoopAhead is the strongest baseline, with $\eta$ of $4.19\times$ on Ouro-2.6B and
$4.12\times$ on Huginn-3.5B. LoopFill improves on these values by
40\% and 25\%. 

At temperature~1.0, LoopFill is the fastest on
five of seven benchmarks on Ouro-2.6B and six of seven on Huginn-3.5B,
with $\eta$ of $3.50\times$ and $2.60\times$. Across the two models, its mean $\eta$ is
$1.49$-$1.52\times$ that of WaveFront, $1.24$-$1.47\times$ that of
LoopSpec, and $1.86$-$2.41\times$ that of SuffixDecoding. LoopFill's $\eta$ exceeds DFlash's by 52\% on Ouro-2.6B and 12\% on
Huginn-3.5B. LoopFill's mean $\eta$ exceeds LoopAhead's by 8\% on Ouro-2.6B
($3.50\times$ vs.\ $3.25\times$) and 10\% on Huginn-3.5B
($2.60\times$ vs.\ $2.37\times$). On Ouro-1.4B, LoopFill also achieves the highest mean decoding speedup over AR across the seven benchmarks, reaching $5.47\times$ under greedy decoding and $3.21\times$ at temperature 1.0 (Table~\ref{tab:main_results_ouro14b}).


Table~\ref{tab:specbench_throughput} compares end-to-end throughput on SpecBench at temperature 1.0. LoopFill achieves the highest throughput on both models at all five concurrency levels. As concurrency increases from $B=1$ to $B=16$, LoopFill's throughput rises, while its throughput relative to AR falls from $4.91\times$ to $2.81\times$ on Ouro-2.6B and from $4.44\times$ to $2.67\times$ on Huginn-3.5B. At $B=16$, LoopFill reaches 189.56 output tokens/s on Ouro-2.6B and 129.18 on Huginn-3.5B, exceeding the next-best method on each model by $1.6\%$ (LoopAhead) and $9.9\%$ (DFlash), respectively. On Ouro-1.4B, LoopFill also leads at all five concurrency levels and achieves $2.50\times$ the throughput of AR at $B=16$ (Table~\ref{tab:ouro14-specbench-throughput}).

\subsection{Ablation Study}

\begin{figure*}[t]
  \centering
  \begin{minipage}[t]{0.25\textwidth}
    \centering
    \includegraphics[width=\linewidth]{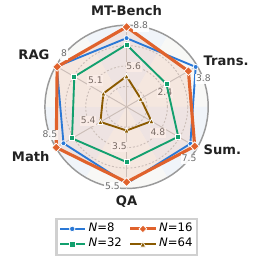}
    \par\smallskip
    \textbf{(a)} Draft budget $N$/ request
  \end{minipage}%
  \begin{minipage}[t]{0.25\textwidth}
    \centering
    \includegraphics[width=\linewidth]{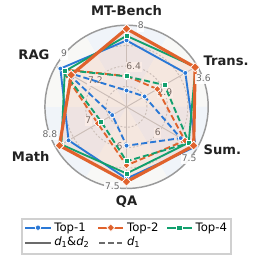}
    \par\smallskip
    \textbf{(b)} Loop Top-$k$ drafts
  \end{minipage}%
  \begin{minipage}[t]{0.25\textwidth}
    \centering
    \includegraphics[width=\linewidth]{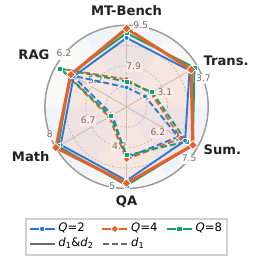}
    \par\smallskip
    \textbf{(c)} Suffix draft budget $Q$
  \end{minipage}%
  \begin{minipage}[t]{0.25\textwidth}
    \centering
    \includegraphics[width=\linewidth]{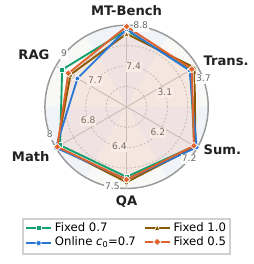}
    \par\smallskip
    \textbf{(d)} Draft tree hyperparameter $c$
  \end{minipage}
  \caption{Ablations of LoopFill on Huginn-3.5B, SpecBench, $B=16$. Each axis is one SpecBench task, including RAG, Multi-turn Conversation (MT-Bench), Translation (Trans.), Summarization(Sum.), QA and Math. The radial value is the throughput (tok/s). Here $d_1$ and $d_2$ are loops 4 and 8 of $T=32$.}
  \label{fig:loopfill-ablation-radar-huginn}
\end{figure*}

\begin{figure}[t]
  \centering
  \includegraphics[width=\columnwidth]{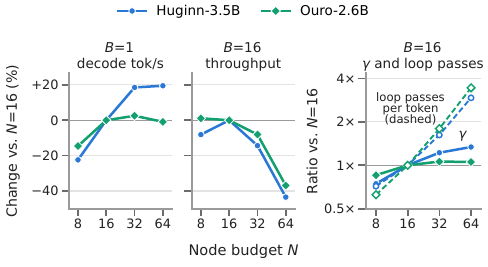}
  \caption{
    \textbf{Effect of the per-request draft-node budget $N$} at two concurrency levels. The left and middle panels show the percentage change from $N=16$ in decoding rate (output tokens/s, $B=1$) and end-to-end throughput ($B=16$), respectively. The right panel shows $\gamma$ (solid) and loop passes per output token (dashed), divided by its value at $N=16, B=16$.}
  \label{fig:loopfill-load-budget}
  \vspace{-2mm}
\end{figure}

\begin{figure}[t]
  \centering
  \includegraphics[width=\columnwidth]{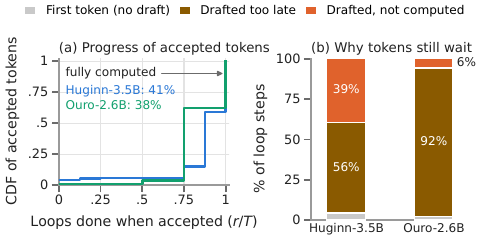}
  \caption{
  \textbf{Accepted tokens are mostly precomputed.}
    \textbf{(a)} Cumulative distribution of $r/T$, where $r$ is the number of loops already completed for the prefix ending in the emitted token.
    \textbf{(b)} Share of decoding loop steps attributed to 
    \emph{First token}: nothing can be drafted before
    decoding starts. \emph{Drafted too late}: the correct token was added to the draft
    tree too late to finish in time. \emph{Drafted, not computed}: it was drafted early
    enough but was pruned.
    }
  \label{fig:loopfill-readiness}
  \vspace{-2mm}
\end{figure}

We explore how several hyperparameters and design choices affect LoopFill’s inference performance on a single L20X GPU.
On Huginn-3.5B with $B=16$ concurrent requests, throughput peaks at
$N=16$ draft nodes per request (Figure~\ref{fig:loopfill-ablation-radar-huginn}(a)). Figures~\ref{fig:loopfill-ablation-radar-huginn}(b) and~\ref{fig:loopfill-ablation-radar-huginn}(c) show the
highest throughput when both shallow drafts propose their two most likely
next tokens ($k=2$) and each suffix lookup adds at most four nodes ($Q=4$).
With $k=2$ and $Q=4$, adding the second draft at $d_2$ raises Huginn's
throughput by 22\% over using $d_1$ alone. Drafts first proposed at $d_2$
account for 11\% of output tokens on Huginn and 5\% on Ouro-2.6B overall,
rising to 26\% and 25\% on translation task (Figures~\ref{fig:loopfill-token-sources} and~\ref{fig:loopfill-token-sources-by-task}).
The weight $c$ scales the initial score of each suffix node relative to
shallow drafts. The online rule starts at $c_0=0.7$ and updates $c$
based on suffix drafts match at verification. Fixing $c$ at 0.5, 0.7,
or 1.0 produces similar throughput (Figure~\ref{fig:loopfill-ablation-radar-huginn}(d)). At $B=1$, $\gamma$ stays
within 3\% of its $c=1$ value in the range from 0.25 to 1.0
(Figure~\ref{fig:loopfill-c-sensitivity}).

We use $N=16$ as a standard setup in Figure~\ref{fig:loopfill-load-budget}.
At $B=1$, a single request can use spare GPU capacity to process more drafts
in parallel and reduce the number of sequential loop steps. Huginn decodes
about 19\% more tokens/ second at $N=32$ or $64$ than at $N=16$.
Increasing $N$ from 32 to 64 raises its $\gamma$ by 9\%, while
decoding time changes only from 30.6 to 30.3\,ms/token. Ouro also improves
little beyond $N=32$. At $B=16$, extra drafts compete with other requests
for GPU time. The rightmost figure in Figure~\ref{fig:loopfill-load-budget} shows \textbf{loop passes per output token}.
Each pass counts one token processed by the looped block, including
tokens later pruned. As $N$ grows, passes per
token increase roughly linearly and the gains in $\gamma$ diminish.
Throughput peaks near $N=16$. Raising $N$ to 64 lowers throughput by 43\%
on Huginn and 37\% on Ouro relative to $N=16$.


Figure~\ref{fig:loopfill-readiness}(a) measures how much computation for the next token has
already been completed when the current token is emitted. When $o_i$
is appended, let $r$ be the number of loops already completed for the
prefix ending in $o_i$. The new root needs $T-r$ more loops to predict
$o_{i+1}$. In 41\% of cases on Huginn and 38\% on Ouro, $r=T$, so the
next token can be emitted in the same loop step. The most common value
is $r=T-d_1$ on both models. This is the progress expected for a child
that starts as soon as its parent reaches the first shallow drafting depth.

Figure~\ref{fig:loopfill-readiness}(b) attributes the decoding loop steps. On
Ouro, 92\% of these steps are attributed to correct drafts proposed too
late to finish, and 6\% to correct drafts pruned under the node budget.
On Huginn, the corresponding shares are 56\% and 39\%. The larger share
attributed to pruning on Huginn compared to Ouro is consistent with the effect of raising
$N$ from 16 to 64 at $B=16$. Mean $\gamma$ increases by 34\% on Huginn
and by 5\% on Ouro in Figure~\ref{fig:loopfill-load-budget}.

%% file: sections/05_relatedwork.tex
\section{Related Work}

\paragraph{Parallel decoding for looped language models.}
Diffusion forcing overlaps computation across tokens by refining several
hidden states in parallel~\citep{geiping2025efficientparallelsamplersrecurrentdepth}.
WaveFront~\citep{ha2026wavefrontdecodingparallelizedselfspeculative} and
LoopSpec~\citep{cho2026loopspecpipelinedselfspeculativedecoding} use
shallow predictions as drafts and verify them at full depth. Each shallow
prediction supplies a candidate for the next token, so drafting further
ahead requires another pass through the shared block.
LoopAhead~\citep{liu2026loopaheadspeculatingaheadfaster} trains a drafter
to predict several tokens from early-loop states and uses deeper-loop
feedback to decide which branches to advance. LoopFill matches the
current prefix against previously seen text to draft several tokens at
once. These tokens can enter the pipeline together, even before the first
shallow prediction for that prefix is available.

\paragraph{Retrieval-based and hybrid drafting.}
The hybrid variant of
SuffixDecoding~\citep{oliaro2025suffixdecodingextremespeculativedecoding}
uses the score of a suffix draft to decide whether to use it or fall
back to EAGLE-3~\citep{li2025eagle3scalinginferenceacceleration}.
RASD~\citep{quan2025rasdretrievalaugmentedspeculativedecoding} merges retrieved token candidates with a draft
model's predictions before verification. RACER~\citep{zhang2026racerretrievalaugmentedcontextualrapid}
also builds a tree, combining retrieved sequences with candidates
from cached target-model logits. LoopFill updates the draft tree
\textbf{during} verification. A retrieved suffix token can propose its successor
once it reaches a shallow loop, and a shallow draft can trigger a fresh
suffix lookup while its parent continues full depth verification. Once the
full-depth output selects a draft branch, LoopFill retains the work completed along that branch.

%% file: sections/06_conclusion.tex
\section{Conclusion}

LoopFill accelerates looped language model decoding without training. Suffix retrieval lets several drafts enter the looped pipeline together, and shallow-loop predictions extend the draft tree in parallel with verification. Across seven benchmarks, the Huginn and Ouro models achieve mean speedups of $5.13-5.85\times$ over autoregressive decoding. On SpecBench, LoopFill achieves $2.50-2.81\times$ the throughput of autoregressive decoding. Nevertheless, LoopFill has two limitations. First, suffix drafts are less likely to match the model’s output when the next token is less predictable. Second, a larger draft budget can reduce single-request latency but lower throughput when concurrency is large. Future work could explore adaptive methods to sustain low decoding latency and high throughput on workloads with varying output predictability and request concurrency.

%% file: sections/07_appendix.tex
\section{appendix}

\subsection{Runtime Setup}
\label{app:runtime_setup}

Each experiment runs on a single NVIDIA L20X GPU (140\,GiB)
in BF16 precision. To reduce variability, we fix the GPU core
and memory clocks at 1980 and 3201\,MHz, respectively, and set
the power limit to 700\,W. Each worker is pinned to four CPU cores.

\paragraph{Inference.} We implement a PyTorch
engine for looped language models. For throughput evaluation,
the engine uses continuous batching with 128-token chunked
prefill, interleaving prompt processing with decoding and
adding new requests as slots become available.
Active draft nodes at different recurrent depths share computation of the recurrent block.
KV caches are stored separately for each layer, with distinct storage for verified tokens and
draft nodes.
After verification, KV caches along the accepted path are copied
into the verified prefix cache, while surviving draft
nodes remain in their states.

\paragraph{Timing.}
We synchronize the GPU at timing boundaries.
Table \ref{tab:main_results} and \ref{tab:main_results_ouro14b} measures decoding latency after prefilling,
and Table \ref{tab:specbench_throughput} measures end-to-end wall-clock time including prefill and decoding.
Both include CPU scheduling and suffix-tree operations.

\subsection{Evaluation Datasets}
\label{app:evaluation-datasets}

Table~\ref{tab:evaluation-datasets} summarizes the evaluation-set
sizes of the 7 benchmarks in our experiments.
Numbers refer to the evaluation sets and exclude
training splits.

\begin{table*}[h]
  \centering
  \small
  \caption{Evaluation-set sizes of the seven benchmarks.}
  \label{tab:evaluation-datasets}
  \begin{tabular}{lrl}
    \toprule
    Dataset & Evaluation-set size & Domain \\
    \midrule
    GSM8K \cite{gsm8k}      & 1,319 &  math \\
    MATH-500 \cite{math}   & 500   &  math \\
    BBH \cite{BBH}        & 6,511 & reasoning \\
    HumanEval+ \cite{humaneval+} & 164   &  code \\
    MBPP+ \cite{humaneval+}      & 378   &  code \\
    Spider \cite{yu2019spiderlargescalehumanlabeleddataset}      & 1034   &  agentic SQL \\
    SpecBench \cite{specbench}
               & 480   & multi-domain \\
    \bottomrule
  \end{tabular}
\end{table*}

\subsection{Training the LoopAhead and DFlash Baselines}
\label{app:drafter-training}

\paragraph{Data.}
We follow the training recipes in
LoopAhead~\cite{liu2026loopaheadspeculatingaheadfaster} and DFlash~\cite{chen2026dflashblockdiffusionflash}.
We train separate drafters for Ouro-2.6B, Ouro-1.4B and Huginn-3.5B. Table~\ref{tab:drafter-data} summarizes the training dataset.
For LoopAhead, we use 9,025 training samples, for DFlash, we use 50,000 training samples. We remove duplicates in each training set, and then sample prompts with seed 42.
The models (Ouro and Huginn) generate ground-truth responses and are used as training labels.

\begin{table*}[h]
  \centering
  \small
  \caption{Dataset for training. Deduplicated against the evaluation set. }
  \label{tab:drafter-data}
  \begin{tabular}{llr}
    \toprule
    Method & Source & Number of samples  \\
    \midrule
    LoopAhead & GSM8K \cite{gsm8k} & 2,000  \\
              & CodeAlpaca \cite{codealpaca} & 2,000  \\
              & MATH \cite{math} & 2,000  \\
              & Nemotron V1 STEM \cite{NemotronPostTrainingDatasetV1} & 2,025  \\
              & UltraChat \cite{ultrachat} & 1,000  \\
              & \textbf{Total} & \textbf{9,025}  \\
    \midrule
    DFlash & Nemotron V1 math \cite{NemotronPostTrainingDatasetV1} & 20,000  \\
           & Nemotron V1 code \cite{NemotronPostTrainingDatasetV1}  & 10,000  \\
           & CodeAlpaca \cite{codealpaca} & 5,000  \\
           & UltraChat \cite{ultrachat} & 15,000  \\
           & \textbf{Total} & \textbf{50,000}  \\
    \bottomrule
  \end{tabular}
\end{table*}

\paragraph{Configuration.}

\textbf{LoopAhead} drafters for both Ouro and Huginn use the same Transformer backbone,
with 2 layers, a hidden width of 512, 8 attention heads,
and a feed-forward width of 1,536. The total trainable parameters are 11.1M for both Ouro-2.6B and Ouro-1.4B, 17.7M for Huginn-3.5B,
including the backbone and model-specific projections layers.
The difference arises from the projection layers,
with hidden state dimensions of 2,048 for Ouro and 5,280 for Huginn. For \textbf{DFlash}, we use the 5-layer block-diffusion drafter and training objective
implemented in SpecForge \cite{specforge}. The drafter has 277.9M trainable
parameters for Ouro-2.6B, 163.6M for Ouro-1.4B, and 332.6M for Huginn-3.5B.

\paragraph{Training Setup.}
Table~\ref{tab:drafter-optimization} lists the training settings.
Both algorithms use BF16 computation and FP32 master weights,
gradient-norm clipping at 1.0.

\begin{table}[h]
  \centering
  \small
  \setlength{\tabcolsep}{4pt}
  \caption{Training settings for LoopAhead and DFlash}
  \label{tab:drafter-optimization}
  \begin{tabular}{@{}lcc@{}}
    \toprule
    Setting & LoopAhead & DFlash \\
    \midrule
    Epochs & 3 & 6 \\
    Batch size & 64 & 32 \\
    Optimizer & AdamW & AdamW \\
    Adam $(\beta_1,\beta_2)$ & $(0.9,0.95)$ & $(0.9,0.999)$ \\
    Weight decay & 0.01 & 0 \\
    Peak learning rate & $2\times10^{-4}$ & $6\times10^{-4}$ \\
    Warmup fraction & 5\% & 4\% \\
    LR decay & Cosine & Cosine \\
    Minimum LR & $2\times10^{-5}$ & 0 \\
    \bottomrule
  \end{tabular}
\end{table}

\subsection{Additional Experiments}

\begin{figure*}[h]
  \centering
  \begin{minipage}[t]{0.25\textwidth}
    \centering
    \includegraphics[width=\linewidth]{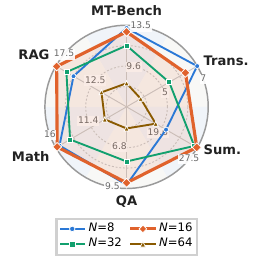}
    \par\smallskip
    \textbf{(a)} Draft budget $N$/ request
  \end{minipage}%
  \begin{minipage}[t]{0.25\textwidth}
    \centering
    \includegraphics[width=\linewidth]{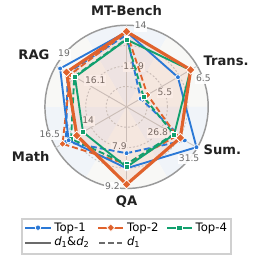}
    \par\smallskip
    \textbf{(b)} Loop Top-$k$ drafts
  \end{minipage}%
  \begin{minipage}[t]{0.25\textwidth}
    \centering
    \includegraphics[width=\linewidth]{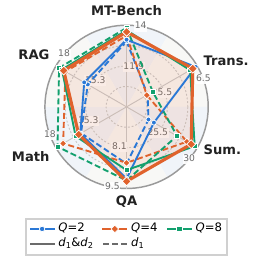}
    \par\smallskip
    \textbf{(c)} Suffix draft budget $Q$
  \end{minipage}%
  \begin{minipage}[t]{0.25\textwidth}
    \centering
    \includegraphics[width=\linewidth]{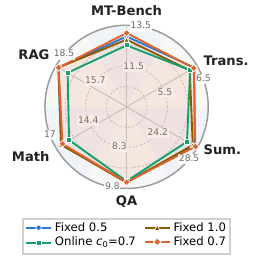}
    \par\smallskip
    \textbf{(d)} Draft tree hyperparameter $c$
  \end{minipage}
  \caption{Ablations of LoopFill on Ouro-2.6B, SpecBench, $B=16$. Here $d_1$ and $d_2$ are loops 1 and 2 of $T=4$.}
  \label{fig:loopfill-ablation-radar-ouro}
\end{figure*}

\begin{figure}[h]
  \centering
  \includegraphics[width=0.45\columnwidth]{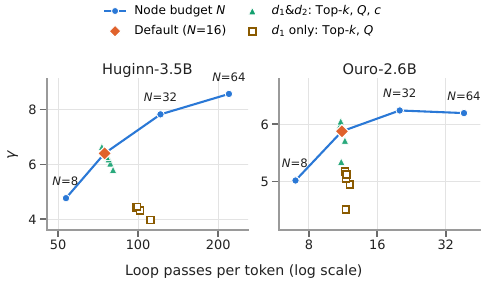}
  \caption{\textbf{Loop passes vs. loop-normalized mean accepted length.}
    The ablation configurations of each model at $B=16$. The $x$-axis is the loop passes per token, the $y$-axis is the loop-normalized mean accepted length $\gamma$. The blue
    line changes the node budget $N$ with everything else default (orange diamond:
    $N=16$, loop-$d_1$\& $d_2$ drafts, Top-2, suffix budget $Q=4$). Green triangles change Top-$k$ (1, 4) or $Q$ (2, 8). Brown squares use
    loop-$d_1$ drafts only (Top-1/2/4 at $Q=4$, $Q=2/8$ at Top-2). Using loop-$d_1$ drafts only lowers $\gamma$ by 29-32\% on Huginn-3.5B and 12-16\% on
    Ouro-2.6B while increasing the compute per token by 30-38\% and 3-6\%.}
  \label{fig:loopfill-ablation-efficiency}
\end{figure}

Figure~\ref{fig:loopfill-ablation-radar-ouro} compares throughput on SpecBench tasks using Ouro-2.6B at
concurrency $B=16$. Increasing the per-request draft budget from $N=16$ to $64$ decreases
throughput on every task. We also change the number of candidates per
shallow draft ($topk$) and the max number of drafts per suffix
retrieval (suffix draft budget $Q$). Figure~\ref{fig:loopfill-ablation-efficiency} calculates loop passes per
output token, including discarded drafts, using greedy decoding
at concurrency $B=16$. On Ouro-2.6B, increasing $N$ from $32$ to $64$ increases loop
passes but barely changes $\gamma$. At $N=16$, removing the second shallow depth
increases loop passes per output token on both models. It also lowers $\gamma$ by 29-32\% on Huginn-3.5B
and 12-16\% on Ouro-2.6B.

\begin{figure*}[t]
  \centering
  \begin{minipage}[t]{0.4\textwidth}
    \centering
    \includegraphics[width=\linewidth]{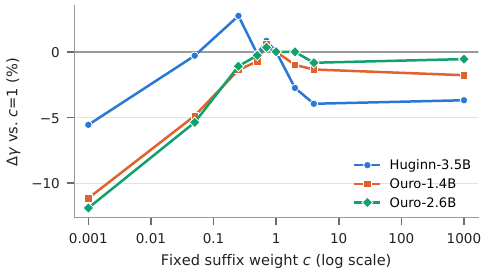}
    \par\smallskip
  \end{minipage}
  \caption{\textbf{Sensitivity of $\gamma$ to suffix weight $c$.}
    Change in the loop-normalized mean accepted length $\gamma$ relative to $c=1$. $B=1$, $N=16$, greedy. $\gamma$ stays within 2\% of $c=1$ for
    $c\in[0.25,1000]$ on both Ouro models and within 4\% for $c\in[0.05,1000]$ on
    Huginn-3.5B. It drops only when suffix-tree drafts are almost ignored: by 5-12\% for Ouro
    at $c\le 0.05$ and by 5.6\% for Huginn-3.5B at $c=0.001$. The suffix-tree draft score $\min(1,cq)$ saturates for large $c$, so $c=4$ and $c=1000$
    behave alike.}
  \label{fig:loopfill-c-sensitivity}
\end{figure*}

\begin{table*}[htbp]
\centering
\caption{
SpecBench end-to-end output throughput at different concurrency levels
for Ouro-1.4B at decoding temperature 1.0.
Each cell gives output tokens/s and the throughput ratio to AR at the same
concurrency. Best results are highlighted in bold.
}
\label{tab:ouro14-specbench-throughput}
\resizebox{0.80\textwidth}{!}{%
\begin{tabular}{llccccc}
\toprule
\textbf{Model} & \textbf{Method}
& $B=1$ & $B=2$ & $B=4$ & $B=8$ & $B=16$ \\
\midrule

\multirow{7}{*}{Ouro-1.4B}
& AR
& 10.77 / 1.00$\times$
& 20.75 / 1.00$\times$
& 40.06 / 1.00$\times$
& 76.25 / 1.00$\times$
& 135.65 / 1.00$\times$ \\

\cmidrule(l){2-7}

& WaveFront
& 26.86 / 2.49$\times$
& 51.00 / 2.46$\times$
& 94.74 / 2.36$\times$
& 168.89 / 2.21$\times$
& 272.37 / 2.01$\times$ \\

& LoopSpec
& 29.77 / 2.76$\times$
& 56.83 / 2.74$\times$
& 105.23 / 2.63$\times$
& 182.51 / 2.39$\times$
& 287.90 / 2.12$\times$ \\

& SuffixDecoding
& 17.55 / 1.63$\times$
& 31.54 / 1.52$\times$
& 61.70 / 1.54$\times$
& 113.74 / 1.49$\times$
& 193.24 / 1.42$\times$ \\

& DFlash
& 31.81 / 2.95$\times$
& 59.24 / 2.85$\times$
& 110.13 / 2.75$\times$
& 193.01 / 2.53$\times$
& 289.28 / 2.13$\times$ \\

& LoopAhead
& 37.48 / 3.48$\times$
& 71.09 / 3.43$\times$
& 124.88 / 3.12$\times$
& 195.35 / 2.56$\times$
& 279.82 / 2.06$\times$ \\

& \textbf{LoopFill}
& \textbf{47.49 / 4.41$\times$}
& \textbf{84.42 / 4.07$\times$}
& \textbf{143.27 / 3.58$\times$}
& \textbf{214.71 / 2.82$\times$}
& \textbf{339.28 / 2.50$\times$} \\

\bottomrule
\end{tabular}%
}
\end{table*}

On Ouro-2.6B at $B=16$, the online update rule and fixed suffix weights
$c\in\{0.5,0.7,1.0\}$ produce similar throughput (Figure~\ref{fig:loopfill-ablation-radar-ouro}(d)).
Figure~\ref{fig:loopfill-c-sensitivity} explores a wider range of suffix weights. On both Ouro-1.4B and Ouro-2.6B, $\gamma$ stays within 2\% of its value at $c=1$ for $c$ from $0.25$ to $1000$. On Huginn-3.5B, it stays within 4\% from
$0.05$ to $1000$. All three models have their lowest $\gamma$ at
$c=0.001$, where suffix drafts have little weight. At large $c$, the
cap in $\min\{1,cq(v)\}$ (Equation~\ref{eq:edge-score}) gives frequent suffix drafts the same initial
weight of one, which helps explain why $\gamma$ levels off.

\begin{figure}[h]
  \centering
  \includegraphics[width=0.5\columnwidth]{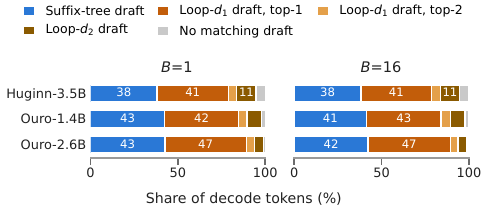}
  \caption{\textbf{Which draft each output token came from.}
    LoopFill on SpecBench. Each bar splits the decode tokens of one model,
    excluding the token produced by prefill. A token is credited to the draft node it matched
    when it was verified: a suffix-tree draft, a loop-$d_1$ draft of rank 1 or 2 in the
    next-token distribution after $d_1$ loops, or a loop-$d_2$ draft first added after $d_2$
    loops ($d_1$ and $d_2$ are loops 1 and 2 of $T=4$ for Ouro and loops 4 and 8 of $T=32$
    for Huginn). Tokens with no matching draft are produced by full-depth verification alone.
    Suffix-tree drafts supply 38-43\% of the tokens and loop-$d_1$/$d_2$ drafts 55-57\%,
    the shares at $B=1$ and $B=16$ differ by at most 1.2 percentage points. Labels give shares
    of at least 10\%.}
  \label{fig:loopfill-token-sources}
\end{figure}

\begin{figure*}[h]
  \centering
  \includegraphics[width=0.9\textwidth]{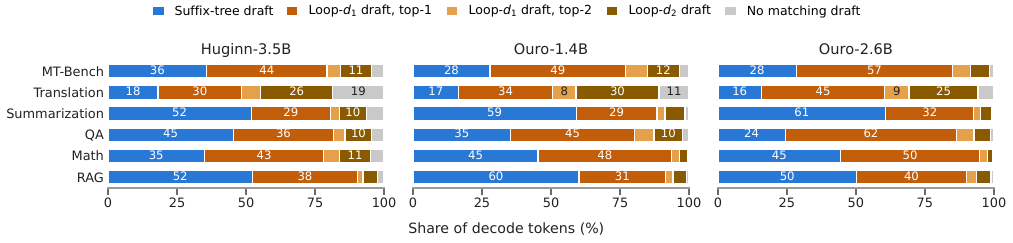}
  \caption{\textbf{Draft sources of the output tokens per SpecBench task at $B=16$.}
    Same runs and definitions as Figure~\ref{fig:loopfill-token-sources}. Suffix-tree drafts
    dominate summarization and RAG (50-61\%), while translation relies on loop-$d_1$/$d_2$
    drafts, including 25-30\% from loop-$d_2$ drafts, and has the most tokens with no
    matching draft (6-19\%). Labels give shares of at least 8\%.}
  \label{fig:loopfill-token-sources-by-task}
\end{figure*}

For each output token after prefill, we record which draft node it matches
at verification, if any (Figure~\ref{fig:loopfill-token-sources}). Across the three models, suffix drafts
account for 38-43\% of SpecBench output tokens and shallow drafts account
for 55-57\%. Increasing concurrency from $B=1$ to $B=16$ changes each
source's share by at most 1.2 percentage points. The shares vary more
across tasks (Figure~\ref{fig:loopfill-token-sources-by-task}). At $B=16$, suffix drafts supply 50-61\% of
the tokens in summarization and retrieval-augmented generation, compared
with 16-18\% in translation. Drafts first proposed at the second shallow
depth $d_2$ account for 25-30\% of translation tokens.

Table~\ref{tab:main_results_ouro14b} extends the single-request latency comparison to Ouro-1.4B.
Under greedy decoding, LoopFill is fastest on all seven benchmarks, with
a mean speedup of $5.47\times$ over AR and a loop-normalized mean accepted
length of $\gamma=6.67$. At temperature $1.0$, the mean speedup falls to
$3.21\times$ and $\gamma$ to $3.91$. LoopFill is fastest on six benchmarks.
On Spider, LoopAhead narrowly leads ($2.94\times$ versus $2.93\times$).
On SpecBench at temperature $1.0$, LoopFill achieves the highest end-to-end
throughput at all five concurrency levels (Table~\ref{tab:ouro14-specbench-throughput}). As $B$ increases from
$1$ to $16$, its throughput rises from $47.49$ to $339.28$ output tokens/s,
while its throughput relative to AR falls from $4.41\times$ to $2.50\times$.

\begin{table*}[h]
\centering
\caption{
Decode latency speedup ($\eta$) and loop-normalized mean accepted length ($\gamma$) of Ouro-1.4B at temperatures 0.0 and 1.0 on seven benchmarks.
Best results are highlighted in \textbf{bold}.
}
\label{tab:main_results_ouro14b}
\vspace{-1mm}
\begingroup
\footnotesize
\renewcommand{\arraystretch}{0.95}
\setlength{\tabcolsep}{3pt}
\begin{adjustbox}{max width=\textwidth}
\begin{tabular}{@{}lcl*{16}{c}@{}}
\toprule
\multirow{2}{*}{Model} & \multirow{2}{*}{Temp.} & \multirow{2}{*}{Method}
& \multicolumn{2}{c}{GSM8K}
& \multicolumn{2}{c}{MATH-500}
& \multicolumn{2}{c}{BBH}
& \multicolumn{2}{c}{HumanEval+}
& \multicolumn{2}{c}{MBPP+}
& \multicolumn{2}{c}{SpecBench}
& \multicolumn{2}{c}{Spider}
& \multicolumn{2}{c}{Mean} \\
& & & $\eta$ & $\gamma$ & $\eta$ & $\gamma$ & $\eta$ & $\gamma$
& $\eta$ & $\gamma$ & $\eta$ & $\gamma$ & $\eta$ & $\gamma$ & $\eta$ & $\gamma$ & $\eta$ & $\gamma$ \\
\midrule
\multirow{14}{*}{Ouro-1.4B} & \multirow{7}{*}{0.0} & AR
& 1.00 & 1.00 & 1.00 & 1.00
& 1.00 & 1.00 & 1.00 & 1.00
& 1.00 & 1.00 & 1.00 & 1.00
& 1.00 & 1.00 & 1.00 & 1.00 \\
& & WaveFront
& 2.59 & 2.82 & 2.96 & 3.20
& 3.01 & 3.22 & 3.37 & 3.57
& 2.97 & 3.27 & 2.47 & 2.66
& 2.46 & 2.68 & 2.83 & 3.06 \\
& & LoopSpec
& 2.94 & 3.25 & 3.15 & 3.50
& 3.22 & 3.52 & 3.42 & 3.75
& 3.17 & 3.56 & 2.84 & 3.09
& 2.78 & 3.07 & 3.07 & 3.39 \\
& & SuffixDecoding
& 1.55 & 1.70 & 1.87 & 2.06
& 2.41 & 2.62 & 2.39 & 2.54
& 1.72 & 1.95 & 1.59 & 1.73
& 1.85 & 2.03 & 1.91 & 2.09 \\
& & DFlash
& 3.01 & 3.45 & 3.33 & 3.82
& 3.53 & 3.98 & 4.44 & 5.17
& 4.38 & 5.18 & 2.59 & 2.97
& 2.54 & 2.88 & 3.40 & 3.92 \\
& & LoopAhead
& 3.70 & 4.72 & 4.41 & 5.74
& 3.75 & 4.79 & 4.79 & 5.97
& 4.25 & 5.48 & 3.12 & 3.88
& 2.85 & 3.52 & 3.84 & 4.87 \\
& & \textbf{LoopFill }
& \textbf{4.54} & \textbf{5.54} & \textbf{5.44} & \textbf{6.84}
& \textbf{6.65} & \textbf{8.07} & \textbf{7.52} & \textbf{8.99}
& \textbf{5.31} & \textbf{6.72} & \textbf{4.31} & \textbf{5.17}
& \textbf{4.55} & \textbf{5.39} & \textbf{5.47} & \textbf{6.67} \\
\cmidrule(l){2-19}
& \multirow{7}{*}{1.0} & AR
& 1.00 & 1.00 & 1.00 & 1.00
& 1.00 & 1.00 & 1.00 & 1.00
& 1.00 & 1.00 & 1.00 & 1.00
& 1.00 & 1.00 & 1.00 & 1.00 \\
& & WaveFront
& 2.25 & 2.37 & 2.32 & 2.50
& 2.18 & 2.36 & 2.40 & 2.69
& 2.08 & 2.43 & 1.72 & 1.96
& 2.66 & 2.20 & 2.23 & 2.36 \\
& & LoopSpec
& 2.29 & 2.59 & 2.43 & 2.66
& 2.34 & 2.55 & 2.55 & 2.85
& 2.28 & 2.58 & 1.80 & 2.10
& 2.39 & 2.31 & 2.30 & 2.52 \\
& & SuffixDecoding
& 1.41 & 1.56 & 1.54 & 1.64
& 1.69 & 1.84 & 1.65 & 1.81
& 1.40 & 1.57 & 1.21 & 1.39
& 1.79 & 1.77 & 1.53 & 1.66 \\
& & DFlash
& 2.32 & 2.75 & 2.42 & 2.80
& 2.28 & 2.69 & 3.03 & 3.48
& 2.64 & 3.06 & 1.73 & 2.12
& 1.87 & 2.15 & 2.33 & 2.72 \\
& & LoopAhead
& 2.72 & 3.74 & 3.09 & 3.89
& 2.64 & 3.20 & 3.13 & 4.21
& 2.98 & 3.86 & 2.10 & 2.78
& \textbf{2.94} & 2.90 & 2.80 & 3.51 \\
& & \textbf{LoopFill}
& \textbf{3.20} & \textbf{3.90} & \textbf{3.37} & \textbf{4.20}
& \textbf{3.27} & \textbf{3.98} & \textbf{3.89} & \textbf{4.80}
& \textbf{3.46} & \textbf{3.99} & \textbf{2.34} & \textbf{2.84}
& 2.93 & \textbf{3.63} & \textbf{3.21} & \textbf{3.91} \\
\bottomrule
\end{tabular}
\end{adjustbox}
\endgroup
\vspace{-1mm}
\end{table*}


%% file: example_paper.bib
@misc{oliaro2025suffixdecodingextremespeculativedecoding,
      title={SuffixDecoding: Extreme Speculative Decoding for Emerging AI Applications}, 
      author={Gabriele Oliaro and Zhihao Jia and Daniel Campos and Aurick Qiao},
      year={2025},
      eprint={2411.04975},
      archivePrefix={arXiv},
      primaryClass={cs.CL},
      url={https://arxiv.org/abs/2411.04975}, 
}

@misc{cho2026loopspecpipelinedselfspeculativedecoding,
      title={LoopSpec: Pipelined Self-Speculative Decoding for Looped Transformers}, 
      author={SangLyul Cho and Langqing Cui and Sehoon Kim and Dongsu Han and Insu Han},
      year={2026},
      eprint={2609.17184},
      archivePrefix={arXiv},
      primaryClass={cs.LG},
      url={https://arxiv.org/abs/2609.17184}, 
}

@misc{ha2026wavefrontdecodingparallelizedselfspeculative,
      title={WaveFront Decoding: Parallelized Self-Speculative Decoding for Looped Language Models}, 
      author={Hyeongju Ha and Jae-Joon Kim},
      year={2026},
      eprint={2609.23033},
      archivePrefix={arXiv},
      primaryClass={cs.LG},
      url={https://arxiv.org/abs/2609.23033}, 
}

@misc{liu2026loopaheadspeculatingaheadfaster,
      title={LoopAhead: Speculating Ahead for Faster Decoding in Looped Language Models}, 
      author={Zixuan Liu and Nan Xue and Rui Bao and Yuhang Xu and Xiaofeng Cao and Jiangchao Yao and Zhiyong Chen},
      year={2026},
      eprint={2610.21594},
      archivePrefix={arXiv},
      primaryClass={cs.AI},
      url={https://arxiv.org/abs/2610.21594}, 
}

@misc{yang2023inferencereferencelosslessacceleration,
      title={Inference with Reference: Lossless Acceleration of Large Language Models}, 
      author={Nan Yang and Tao Ge and Liang Wang and Binxing Jiao and Daxin Jiang and Linjun Yang and Rangan Majumder and Furu Wei},
      year={2023},
      eprint={2304.04487},
      archivePrefix={arXiv},
      primaryClass={cs.CL},
      url={https://arxiv.org/abs/2304.04487}, 
}

@misc{saxena2023prompt,
    title = {Prompt Lookup Decoding},
    author = {Apoorv Saxena},
    year = {2023},
    month = {November},
    url = {https://github.com/apoorvumang/prompt-lookup-decoding/}
}

@inproceedings{ou-etal-2024-lossless,
    title = "Lossless Acceleration of Large Language Model via Adaptive N-gram Parallel Decoding",
    author = "Ou, Jie  and
      Chen, Yueming  and
      Tian, Wenhong",
    editor = "Yang, Yi  and
      Davani, Aida  and
      Sil, Avi  and
      Kumar, Anoop",
    booktitle = "Proceedings of the 2024 Conference of the {N}orth {A}merican Chapter of the {A}ssociation for {C}omputational {L}inguistics: Human Language Technologies (Volume 6: Industry Track)",
    month = jun,
    year = "2024",
    address = "Mexico City, Mexico",
    publisher = "Association for Computational Linguistics",
    url = "https://aclanthology.org/2024.naacl-industry.2/",
    doi = "10.18653/v1/2024.naacl-industry.2",
    pages = "10--22"
}

@inproceedings{luo-etal-2025-turning,
    title = "Turning Trash into Treasure: Accelerating Inference of Large Language Models with Token Recycling",
    author = "Luo, Xianzhen  and
      Wang, Yixuan  and
      Zhu, Qingfu  and
      Zhang, Zhiming  and
      Zhang, Xuanyu  and
      Yang, Qing  and
      Xu, Dongliang",
    editor = "Che, Wanxiang  and
      Nabende, Joyce  and
      Shutova, Ekaterina  and
      Pilehvar, Mohammad Taher",
    booktitle = "Proceedings of the 63rd Annual Meeting of the {A}ssociation for {C}omputational {L}inguistics (Volume 1: Long Papers)",
    month = jul,
    year = "2025",
    address = "Vienna, Austria",
    publisher = "Association for Computational Linguistics",
    url = "https://aclanthology.org/2025.acl-long.338/",
    doi = "10.18653/v1/2025.acl-long.338",
    pages = "6816--6831",
    ISBN = "979-8-89176-251-0"
}

@misc{zhu2026scalinglatentreasoninglooped,
      title={Scaling Latent Reasoning via Looped Language Models}, 
      author={Rui-Jie Zhu and Zixuan Wang and Kai Hua and Tianyu Zhang and Ziniu Li and Haoran Que and Boyi Wei and Zixin Wen and Fan Yin and He Xing and Lu Li and Jiajun Shi and Kaijing Ma and Shanda Li and Taylor Kergan and Andrew Smith and Xingwei Qu and Mude Hui and Bohong Wu and Qiyang Min and Hongzhi Huang and Xun Zhou and Wei Ye and Jiaheng Liu and Jian Yang and Yunfeng Shi and Chenghua Lin and Enduo Zhao and Tianle Cai and Ge Zhang and Wenhao Huang and Yoshua Bengio and Jason Eshraghian},
      year={2026},
      eprint={2510.25741},
      archivePrefix={arXiv},
      primaryClass={cs.CL},
      url={https://arxiv.org/abs/2510.25741}, 
}

@misc{geiping2025scalingtesttimecomputelatent,
      title={Scaling up Test-Time Compute with Latent Reasoning: A Recurrent Depth Approach}, 
      author={Jonas Geiping and Sean McLeish and Neel Jain and John Kirchenbauer and Siddharth Singh and Brian R. Bartoldson and Bhavya Kailkhura and Abhinav Bhatele and Tom Goldstein},
      year={2025},
      eprint={2502.05171},
      archivePrefix={arXiv},
      primaryClass={cs.LG},
      url={https://arxiv.org/abs/2502.05171}, 
}

@inproceedings{dehghani2019universaltransformers,
  title     = {Universal Transformers},
  author    = {Dehghani, Mostafa and Gouws, Stephan and Vinyals, Oriol and Uszkoreit, Jakob and Kaiser, {\L}ukasz},
  booktitle = {International Conference on Learning Representations},
  year      = {2019},
  url       = {https://arxiv.org/abs/1807.03819}
}

@inproceedings{bae2025mixtureofrecursionslearningdynamicrecursive,
  title     = {Mixture-of-Recursions: Learning Dynamic Recursive Depths for Adaptive Token-Level Computation},
  author    = {Bae, Sangmin and Kim, Yujin and Bayat, Reza and Kim, Sungnyun and Ha, Jiyoun and Schuster, Tal and Fisch, Adam and Harutyunyan, Hrayr and Ji, Ziwei and Courville, Aaron and Yun, Se-Young},
  booktitle = {Advances in Neural Information Processing Systems},
  volume    = {38},
  year      = {2025},
  url       = {https://arxiv.org/abs/2507.10524}
}

@inproceedings{saunshi2025reasoninglatentthoughtspower,
  title     = {Reasoning with Latent Thoughts: On the Power of Looped Transformers},
  author    = {Saunshi, Nikunj and Dikkala, Nishanth and Li, Zhiyuan and Kumar, Sanjiv and Reddi, Sashank J.},
  booktitle = {International Conference on Learning Representations},
  year      = {2025},
  url       = {https://arxiv.org/abs/2502.17416}
}

@misc{pope2022efficientlyscalingtransformerinference,
      title={Efficiently Scaling Transformer Inference}, 
      author={Reiner Pope and Sholto Douglas and Aakanksha Chowdhery and Jacob Devlin and James Bradbury and Anselm Levskaya and Jonathan Heek and Kefan Xiao and Shivani Agrawal and Jeff Dean},
      year={2022},
      eprint={2211.05102},
      archivePrefix={arXiv},
      primaryClass={cs.LG},
      url={https://arxiv.org/abs/2211.05102}, 
}

@misc{chen2026dflashblockdiffusionflash,
      title={DFlash: Block Diffusion for Flash Speculative Decoding}, 
      author={Jian Chen and Yesheng Liang and Zhijian Liu},
      year={2026},
      eprint={2602.06036},
      archivePrefix={arXiv},
      primaryClass={cs.CL},
      url={https://arxiv.org/abs/2602.06036}, 
}

@misc{gsm8k,
      title={Training Verifiers to Solve Math Word Problems}, 
      author={Karl Cobbe and Vineet Kosaraju and Mohammad Bavarian and Mark Chen and Heewoo Jun and Lukasz Kaiser and Matthias Plappert and Jerry Tworek and Jacob Hilton and Reiichiro Nakano and Christopher Hesse and John Schulman},
      year={2021},
      eprint={2110.14168},
      archivePrefix={arXiv},
      primaryClass={cs.LG},
      url={https://arxiv.org/abs/2110.14168}, 
}

@misc{codealpaca,
  author = {Sahil Chaudhary},
  title = {Code Alpaca: An Instruction-following LLaMA model for code generation},
  year = {2023},
  publisher = {GitHub},
  journal = {GitHub repository},
  howpublished = {\url{https://github.com/sahil280114/codealpaca}},
}

@misc{math,
      title={Measuring Mathematical Problem Solving With the MATH Dataset}, 
      author={Dan Hendrycks and Collin Burns and Saurav Kadavath and Akul Arora and Steven Basart and Eric Tang and Dawn Song and Jacob Steinhardt},
      year={2021},
      eprint={2103.03874},
      archivePrefix={arXiv},
      primaryClass={cs.LG},
      url={https://arxiv.org/abs/2103.03874}, 
}

@misc{ultrachat,
      title={Enhancing Chat Language Models by Scaling High-quality Instructional Conversations}, 
      author={Ning Ding and Yulin Chen and Bokai Xu and Yujia Qin and Zhi Zheng and Shengding Hu and Zhiyuan Liu and Maosong Sun and Bowen Zhou},
      year={2023},
      eprint={2305.14233},
      archivePrefix={arXiv},
      primaryClass={cs.CL},
      url={https://arxiv.org/abs/2305.14233}, 
}

@software{NemotronPostTrainingDatasetV1,
      author = {Nathawani, Dhruv and Gitman, Igor and Majumdar, Somshubra and Bakhturina, Evelina and Sunil Mahabaleshwarkar, Ameya and and Zhang, Jian and Polak Scowcroft, Jane},
      title = {{Nemotron-Post-Training-Dataset-v1}},
      version = {1.0},
      publisher = {{NVIDIA}},
      year = {2025}, month = July,
      url = {https://huggingface.co/datasets/nvidia/Nemotron-Post-Training-Dataset-v1}
}

@misc{specforge,
      title={SpecForge: A Flexible and Efficient Open-Source Training Framework for Speculative Decoding}, 
      author={Shenggui Li and Chao Wang and Yikai Zhu and Yubo Wang and Fan Yin and Shuai Shi and Yefei Chen and Xiaomin Dong and Qiaoling Chen and Jin Pan and Ji Li and Laixin Xie and Yineng Zhang and Lei Yu and Yonggang Wen and Ivor Tsang and Tianwei Zhang},
      year={2026},
      eprint={2603.18567},
      archivePrefix={arXiv},
      primaryClass={cs.LG},
      url={https://arxiv.org/abs/2603.18567}, 
}

@misc{li2025eagle3scalinginferenceacceleration,
      title={EAGLE-3: Scaling up Inference Acceleration of Large Language Models via Training-Time Test}, 
      author={Yuhui Li and Fangyun Wei and Chao Zhang and Hongyang Zhang},
      year={2025},
      eprint={2503.01840},
      archivePrefix={arXiv},
      primaryClass={cs.CL},
      url={https://arxiv.org/abs/2503.01840}, 
}

@misc{BBH,
      title={Challenging BIG-Bench Tasks and Whether Chain-of-Thought Can Solve Them}, 
      author={Mirac Suzgun and Nathan Scales and Nathanael Schärli and Sebastian Gehrmann and Yi Tay and Hyung Won Chung and Aakanksha Chowdhery and Quoc V. Le and Ed H. Chi and Denny Zhou and Jason Wei},
      year={2022},
      eprint={2210.09261},
      archivePrefix={arXiv},
      primaryClass={cs.CL},
      url={https://arxiv.org/abs/2210.09261}, 
}

@misc{humaneval+,
      title={Is Your Code Generated by ChatGPT Really Correct? Rigorous Evaluation of Large Language Models for Code Generation}, 
      author={Jiawei Liu and Chunqiu Steven Xia and Yuyao Wang and Lingming Zhang},
      year={2023},
      eprint={2305.01210},
      archivePrefix={arXiv},
      primaryClass={cs.SE},
      url={https://arxiv.org/abs/2305.01210}, 
}

@inproceedings{specbench,
    title = "Unlocking Efficiency in Large Language Model Inference: A Comprehensive Survey of Speculative Decoding",
    author = "Xia, Heming and Yang, Zhe and Dong, Qingxiu and Wang, Peiyi and Li, Yongqi  and Ge, Tao and Liu, Tianyu and Li, Wenjie and Sui, Zhifang",
    editor = "Ku, Lun-Wei and Martins, Andre and Srikumar, Vivek",
    booktitle = "Findings of the Association for Computational Linguistics ACL 2024",
    month = aug,
    year = "2024",
    address = "Bangkok, Thailand and virtual meeting",
    publisher = "Association for Computational Linguistics",
    url = "https://aclanthology.org/2024.findings-acl.456",
    doi = "10.18653/v1/2024.findings-acl.456",
    pages = "7655--7671",
}

@misc{yu2019spiderlargescalehumanlabeleddataset,
      title={Spider: A Large-Scale Human-Labeled Dataset for Complex and Cross-Domain Semantic Parsing and Text-to-SQL Task}, 
      author={Tao Yu and Rui Zhang and Kai Yang and Michihiro Yasunaga and Dongxu Wang and Zifan Li and James Ma and Irene Li and Qingning Yao and Shanelle Roman and Zilin Zhang and Dragomir Radev},
      year={2019},
      eprint={1809.08887},
      archivePrefix={arXiv},
      primaryClass={cs.CL},
      url={https://arxiv.org/abs/1809.08887}, 
}

@misc{hooper2024speedspeculativepipelinedexecution,
      title={SPEED: Speculative Pipelined Execution for Efficient Decoding}, 
      author={Coleman Hooper and Sehoon Kim and Hiva Mohammadzadeh and Hasan Genc and Kurt Keutzer and Amir Gholami and Sophia Shao},
      year={2024},
      eprint={2310.12072},
      archivePrefix={arXiv},
      primaryClass={cs.CL},
      url={https://arxiv.org/abs/2310.12072}, 
}

@misc{geiping2025efficientparallelsamplersrecurrentdepth,
      title={Efficient Parallel Samplers for Recurrent-Depth Models and Their Connection to Diffusion Language Models}, 
      author={Jonas Geiping and Xinyu Yang and Guinan Su},
      year={2025},
      eprint={2510.14961},
      archivePrefix={arXiv},
      primaryClass={cs.LG},
      url={https://arxiv.org/abs/2510.14961}, 
}

@misc{quan2025rasdretrievalaugmentedspeculativedecoding,
      title={RASD: Retrieval-Augmented Speculative Decoding}, 
      author={Guofeng Quan and Wenfeng Feng and Chuzhan Hao and Guochao Jiang and Yuewei Zhang and Hao Wang},
      year={2025},
      eprint={2503.03434},
      archivePrefix={arXiv},
      primaryClass={cs.CL},
      url={https://arxiv.org/abs/2503.03434}, 
}

@misc{zhang2026racerretrievalaugmentedcontextualrapid,
      title={RACER: Retrieval-Augmented Contextual Rapid Speculative Decoding}, 
      author={Zihong Zhang and Zuchao Li and Lefei Zhang and Ping Wang and Hai Zhao},
      year={2026},
      eprint={2604.14885},
      archivePrefix={arXiv},
      primaryClass={cs.CL},
      url={https://arxiv.org/abs/2604.14885}, 
}
